\documentclass{article} 
\usepackage{iclr2021_conference,times}

\usepackage{todonotes}

\usepackage{amsmath,amsfonts,bm}

\def\eqref#1{equation~\ref{#1}}

\def\1{\bm{1}}

\DeclareMathAlphabet{\mathsfit}{\encodingdefault}{\sfdefault}{m}{sl}
\SetMathAlphabet{\mathsfit}{bold}{\encodingdefault}{\sfdefault}{bx}{n}

\newcommand{\E}{\mathbb{E}}

\DeclareMathOperator*{\argmin}{arg\,min}

\usepackage{hyperref}
\usepackage{url}
\usepackage{graphicx}
\usepackage{blindtext}
\usepackage{tcolorbox}
\usepackage{subcaption}
\usepackage{multirow}
\usepackage{mathtools}
\usepackage{amsmath,psfrag,pstcol,amssymb}
\usepackage{float}
\usepackage{algorithm}
\usepackage{algorithmic}
\newcommand{\INPUT}{\textbf{Inputs: }}
\newcommand{\OUTPUT}{\textbf{Outputs: }}
\algsetup{
  linenodelimiter = {}
}
\title{$\delta$-CLUE: Diverse Sets of Explanations for\\ Uncertainty Estimates}

\author{%
  Dan Ley\\
  University of Cambridge\\
  \texttt{dwl36@cantab.ac.uk} \\
   \And
   Umang Bhatt \\
   University of Cambridge \\
   \texttt{usb20@cam.ac.uk} \\
   \And
   Adrian Weller \\
   University of Cambridge\\The Alan Turing Institute \\
   \texttt{aw665@cam.ac.uk} \\
}

\iclrfinalcopy 
\begin{document}

\maketitle

\begin{abstract}
To interpret uncertainty estimates from differentiable probabilistic models, recent work has proposed generating Counterfactual Latent Uncertainty Explanations (CLUEs). However, for a single input, such approaches could output a variety of explanations due to the lack of constraints placed on the explanation. Here we augment the original CLUE approach, to provide what we call $\delta$-CLUE. CLUE indicates \emph{one} way to change an input, while remaining on the data manifold, such that the model becomes more confident about its prediction. We instead return a \emph{set} of plausible CLUEs: multiple, diverse inputs that are within a $\delta$ ball of the original input in latent space, all yielding confident predictions.
\end{abstract}

\section{Introduction}
For models that provide uncertainty estimates alongside their predictions, explaining the source of this uncertainty reveals important information.
\citet{antoran2021getting} propose a method for finding an explanation of a model's predictive uncertainty of a given input by searching in the latent space of an auxiliary deep generative model (DGM): they identify a single possible change to the input, while keeping it in distribution, such that the model becomes more certain in its prediction. 
Termed CLUE (Counterfactual Latent Uncertainty Explanation), this method is effective for generating 
plausible changes to an input that reduce uncertainty. These changes are distinct from adversarial examples, which instead find nearby points that change the label~\citep{goodfellow2014explaining}. However, there are limitations to CLUE, including 
the lack of a framework to deal with a potential diverse set of plausible explanations~\citep{russell2019efficient}, despite proposing methods to generate them.

CLUE introduces a latent variable DGM: $p_\theta(\mathbf{x}) = \int(p_\theta(\mathbf{x}|\mathbf{z})p(\mathbf{z})d\mathbf{z}$, with encoder $q_\phi(\mathbf{z}|\mathbf{x})$. The predictive mean of the DGM is $\E_{p_\theta(\mathbf{x}|\mathbf{z})}[\mathbf{x}]=\mu_\theta(\mathbf{x}|\mathbf{z})$ and of the encoder is $\E_{q_\phi(\mathbf{z}|\mathbf{x})}[\mathbf{z}]=\mu_\phi(\mathbf{z}|\mathbf{x})$ respectively. $\mathcal{H}$ refers to any differentiable uncertainty estimate of a prediction $\mathbf{y}$. CLUE minimises:
\vspace{-8pt}
\begin{equation}\label{eq:1}\mathcal{L}(\mathbf{z})=\mathcal{H}\left(\mathbf{y}|\mu_{\theta}(\mathbf{x}|\mathbf{z})\right)+d\left(\mu_{\theta}(\mathbf{x}|\mathbf{z}), \mathbf{x}_{0}\right), \end{equation}
\vspace{-12pt}
\begin{equation}\label{eq:2}\text{to yield  } \quad \mathbf{x}_{\mathrm{CLUE}}=\mu_{\theta}\left(\mathbf{x}|\mathbf{z}_{\mathrm{CLUE}}\right)\ \text{ where }\ \ \mathbf{z}_{\mathrm{CLUE}}=\argmin_{\mathbf{z}} \mathcal{L}(\mathbf{z}).\end{equation}
The pairwise distance metric takes the form $d(\mathbf{x}, \mathbf{x}_0) = \lambda_xd_x(\mathbf{x}, \mathbf{x}_0) + \lambda_yd_y(f(\mathbf{x}), f(\mathbf{x}_0))$, where $f(\mathbf{x})$ is the model's mapping from an input $x$ to a label, thus encouraging similarity between uncertain points and CLUEs in both input and prediction space.



\begin{figure}[b]
    \centering
    \includegraphics[scale=0.4]{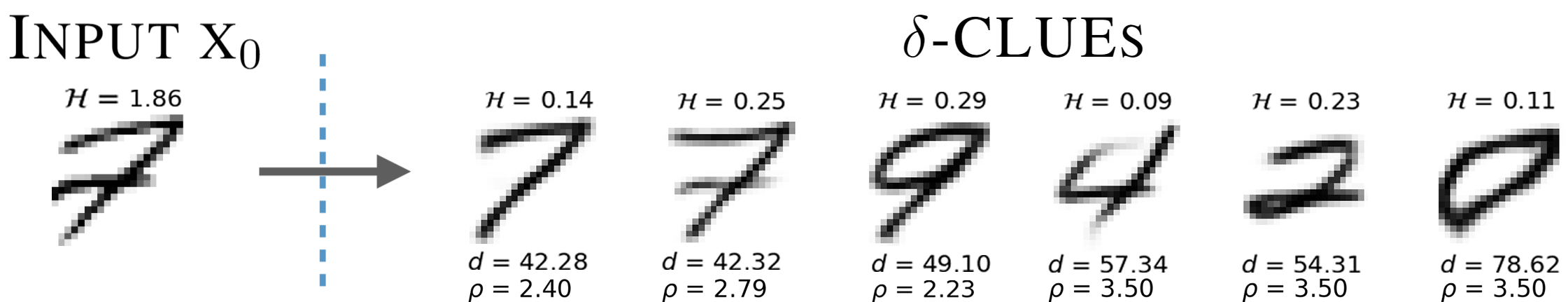}
    \caption{\small We produce a \textbf{diverse set} of candidate explanations that show how to reduce predictive uncertainty while still remaining close to $x_0$ in both input and latent space ($\mathcal{H}$ is uncertainty, $d$ is input space distance, $\rho$ is latent space distance). We see that the left image might easily be resolved into a confident 7 or 9.}
    \label{fig:digits}
\end{figure}

In this paper, we tackle the problem of finding multiple, diverse CLUEs. Providing practitioners with many explanations for why their input was uncertain can be helpful if, for instance, they are not in control of the recourse suggestions proposed by the algorithm; advising someone to change their age is less actionable than advising them to change a mutable characteristic~\citep{poyiadzi2020face}.

\section{Methodology}

\begin{figure}[H]
    \centering
    \includegraphics[scale=0.43]{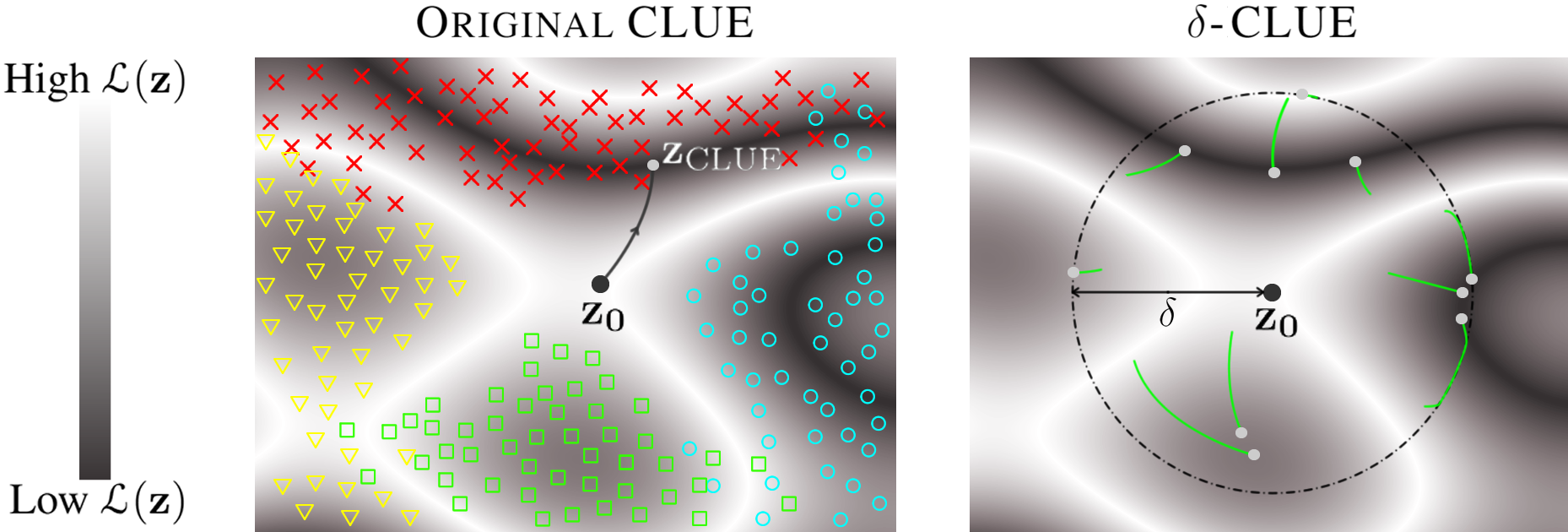}
    \caption{\small Conceptual colour map of objective function $\mathcal{L}(z)$ with $\mathbf{z}_0$ located in high cost region. Left: Gradient descent to region of low cost (original CLUE algorithm). Training points are shown in colour. Right: Gradient descent constrained to $\delta$-ball at every step. Diverse starting points yield diverse local minima. White circles indicate 
    CLUEs found.}
    \label{fig:DeltaFinal}
\end{figure}

We propose to modify the original method to generate a set of solutions that are all within a specified distance $\delta$ of $\mathbf{z}_0=\mu_\phi(\mathbf{z}|\mathbf{x_0})$ in latent space: $\mathbf{z}_0$ is the latent space representation of the uncertain input $\mathbf{x}_0$ being explained. We achieve multiplicity by initialising the search in different areas of latent space using varied initialisation methods $\mathcal{S}_i$. Experiments are performed on the MNIST dataset~\citep{lecun1998mnist}, where finding diverse CLUEs amounts to maximising the number of class labels we converge to in the search. Figure~\ref{fig:DeltaFinal} contrasts the original and proposed objectives.

In the original CLUE objective, the DGM and neural networks used are VAEs~\citep{ivanov2018variational} and BNNs~\citep{gal2016uncertainty} respectively. The uncertainty of the BNN for a point is given by the entropy of the  posterior over the class labels; we use the same measure. 
The hyperparameters ($\lambda_x, \lambda_y$) control the trade-off between producing low uncertainty CLUEs and CLUEs which are close to the original inputs. To encourage sparse explanations, we take $d_x(\mathbf{x}, \mathbf{x}_0) = \|\mathbf{x}-\mathbf{x}_0\|_1$: see Appendix \ref{appendix:distancemetrics} for trade-offs. Figure~\ref{fig:DeltaFinal} (left) shows a conceptual path taken by this optimisation.
In our proposed $\delta$-CLUE method, the loss function 
is the same as in Eq~\ref{eq:1}, with the additional $\delta$ requirement as:
\begin{equation}\mathbf{x}_{\delta-\mathrm{CLUE}}=\mu_{\theta}\left(\mathbf{x}|\mathbf{z}_{\delta-\mathrm{CLUE}}\right)\ \text{ where }\ \ \mathbf{z}_{\delta-\mathrm{CLUE}}=\argmin_{\mathbf{z}:\ \rho(\mathbf{z}, \mathbf{z}_0)\leq\delta} \mathcal{L}(\mathbf{z})\ \text{ and }\ \ \mathbf{z_0}=\mu_\phi(\mathbf{z}|\mathbf{x_0}).\end{equation}
We choose $\rho(\mathbf{z}, \mathbf{z}_0)=\|\mathbf{z}-\mathbf{z}_0\|_2$ (the Euclidean $\ell_2$ norm) in this paper, as shown in the 2D depiction in Figure \ref{fig:DeltaFinal}. We first set $\lambda_x=\lambda_y=0$ to explore solely the uncertainty landscape, given that the size of the $\delta$-ball removes the strict need for the distance component in $\mathcal{L}(z)$ and grants control over the locality of solutions, before trialling $\lambda_x\approx0.03$.
The $\delta$ constraint can be applied either throughout each stage of the optimisation as in Projected Gradient Descent~\citep{boyd2004convex} (Figure~\ref{fig:DeltaFinal}, right) or post optimisation (Appendix~\ref{appendix:constrainedvsunconstrained}). The optimal $\delta$ value(s) can be determined through experimentation (Figure \ref{fig:uncertdistmin}), although Appendix \ref{appendix:constrainedvsunconstrained} discusses other potential methods.


For each uncertain input $x_0$, we exploit the non-convexity of CLUE's objective to generate diverse $\delta$-CLUEs by initialising gradient descents in different regions of latent space to converge to different local minima (Figure \ref{fig:DeltaFinal}). We propose multiple initialisation schemes, $\mathcal{S}_i$; some may randomly initialise within the $\delta$-ball, while others could use training data or class boundaries to determine starting points (shown in dark blue in Figure~\ref{fig:SearchStrats}). We describe the $\delta$-CLUE method in Algorithm 1.

\section{Experiments Groups}

We perform constrained optimisation during gradient descent (Figure \ref{fig:DeltaFinal}, right). Appendix~\ref{appendix:constrainedvsunconstrained} provides justification for this decision. In our experiments, we search in the latent space of a VAE to generate $\delta$-CLUEs for the $8$ most uncertain digits in the MNIST test set, according to our trained BNN.

We trial this over \textbf{a)} a range of several $\delta$ values from $0.5$ to $3.5$, \textbf{b)} two latent space loss functions: \textbf{Uncertainty} $\mathcal{L}_{\mathcal{H}}=\mathcal{H}$ and \textbf{Distance} $\mathcal{L}_{\mathcal{H}+d}=\mathcal{H}+d$ and \textbf{c)} two initialisation schemes as depicted in Figure~\ref{fig:SearchStrats}. Initialisation scheme $\mathcal{S}_1$ picks a random direction at a uniform random radius within the delta ball, while the other scheme $\mathcal{S}_2$ is along paths determined by the nearest neighbours (\textbf{NN}) for each class in the training data.  We label these experiment variants as: \textbf{Uncertainty Random}: [$\mathcal{L}_{\mathcal{H}}$, $\mathcal{S}_1$], \textbf{Uncertainty NN}: [$\mathcal{L}_{\mathcal{H}}$, $\mathcal{S}_2$], \textbf{Distance Random}: [$\mathcal{L}_{\mathcal{H}+d}$, $\mathcal{S}_1$] and \textbf{Distance NN}: [$\mathcal{L}_{\mathcal{H}+d}$, $\mathcal{S}_2$].

\begin{minipage}{.55\textwidth}
\begin{algorithm}[H]
 \INPUT radius of search $\delta$, number of explanations $n$, initialisation scheme $\mathcal{S}_i$, original datapoint $\mathbf{x}_0$, input space distance function $d$, latent space distance function $\rho$, BNN uncertainty estimator $\mathcal{H}$, DGM decoder $\mu_\theta(\cdot)$, DGM encoder $\mu_\phi(\cdot)$
 \begin{algorithmic}[1]
 \STATE Set $\delta$-ball centre of $\mathbf{z}_0=\mu_\phi(\mathbf{z}|\mathbf{x}_0)$;
 \FORALL{explanations $i\leq n$}
     \STATE Set initial value of $\mathbf{z}_i = \mathcal{S}(\mathbf{z}_0, \delta, i, n)$;
     \WHILE{\textit{loss $\mathit{\mathcal{L}}$ is not converged}}
         \STATE Decode: $\mathbf{x} = \mu_\theta(\mathbf{x}|\mathbf{z}_i)$;
         \STATE Use BNN to obtain $\mathcal{H}(\mathbf{y}|\mathbf{x})$;
         \STATE $\mathcal{L}=\mathcal{H}(\mathbf{y}|\mathbf{x})+d(\mathbf{x}, \mathbf{x}_0)$;
         \STATE Update $\mathbf{z}_i$ with $\nabla_\mathbf{z}\mathcal{L}$;
         \IF{$\rho(\mathbf{z}_i, \mathbf{z}_0)>\delta$}
            \STATE Project $\mathbf{z}_i$ onto the surface of the $\delta$-ball as $\mathbf{z}_i = \delta\times\frac{\mathbf{z}_i-\mathbf{z}_0}{\rho(\mathbf{z}_i,\mathbf{z}_0)}$;
         \ENDIF
     \ENDWHILE
     \STATE Decode explanation: $\mathbf{x}_{\delta-\mathrm{CLUE}}=\mu_\theta(\mathbf{x}|\mathbf{z}_i)$;
     \STATE Accept if $\mathcal{H}(\mathbf{y}|\mathbf{x}_{\delta-\mathrm{CLUE}}) < \mathcal{H}_{\mathrm{threshold}}$;
 \ENDFOR
 \end{algorithmic}
 \OUTPUT A set of $m\leq n$ $\delta$-CLUEs $\mathbf{x}_{\delta-\mathrm{CLUE}}$
\caption{: $\delta$-CLUE}
\end{algorithm}
\end{minipage}
\hspace{.05\textwidth}
\begin{minipage}{.4\textwidth}
\centering
\includegraphics[scale=0.48]{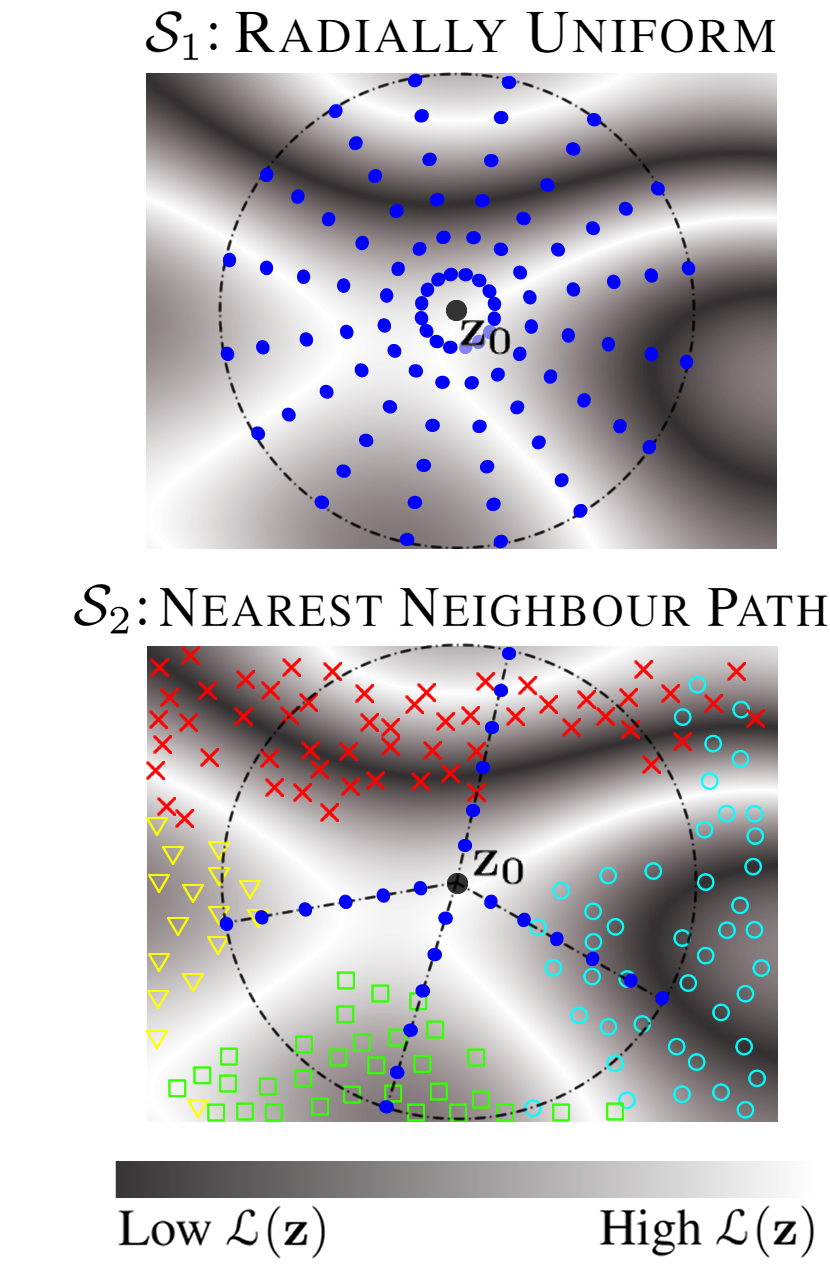}
\captionof{figure}{\small Two possible initialisation schemes $\mathcal{S}_i$ to yield diverse minima. One is random, the other deterministic. Details are provided in Appendix \ref{appendix:schemes}.}
\label{fig:SearchStrats}
\end{minipage}

\begin{figure}[H]
\begin{subfigure}{0.5\textwidth}
    \centering
    \includegraphics[scale=0.29]{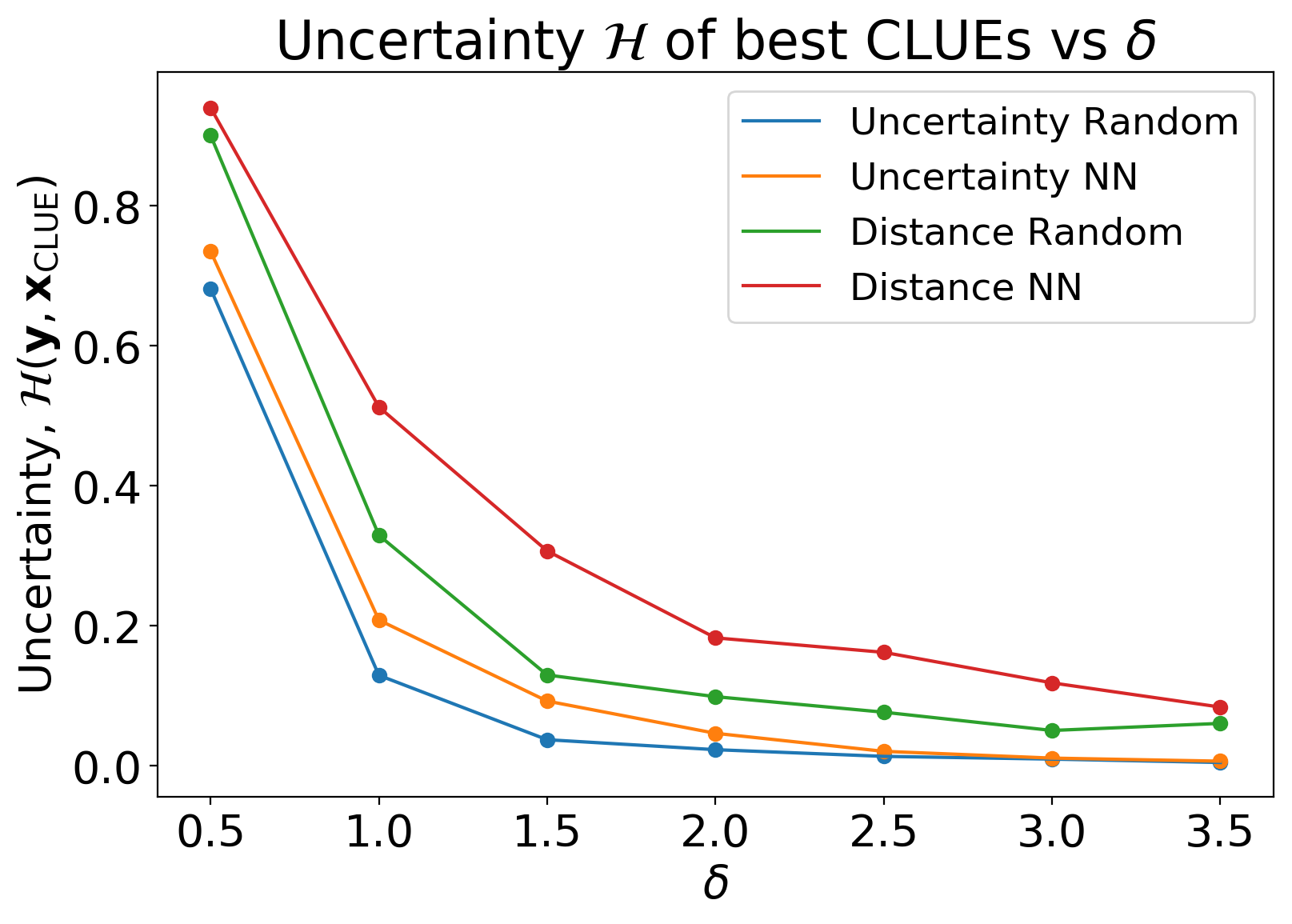}
\end{subfigure}
\begin{subfigure}{0.5\textwidth}
    \centering
    \includegraphics[scale=0.29]{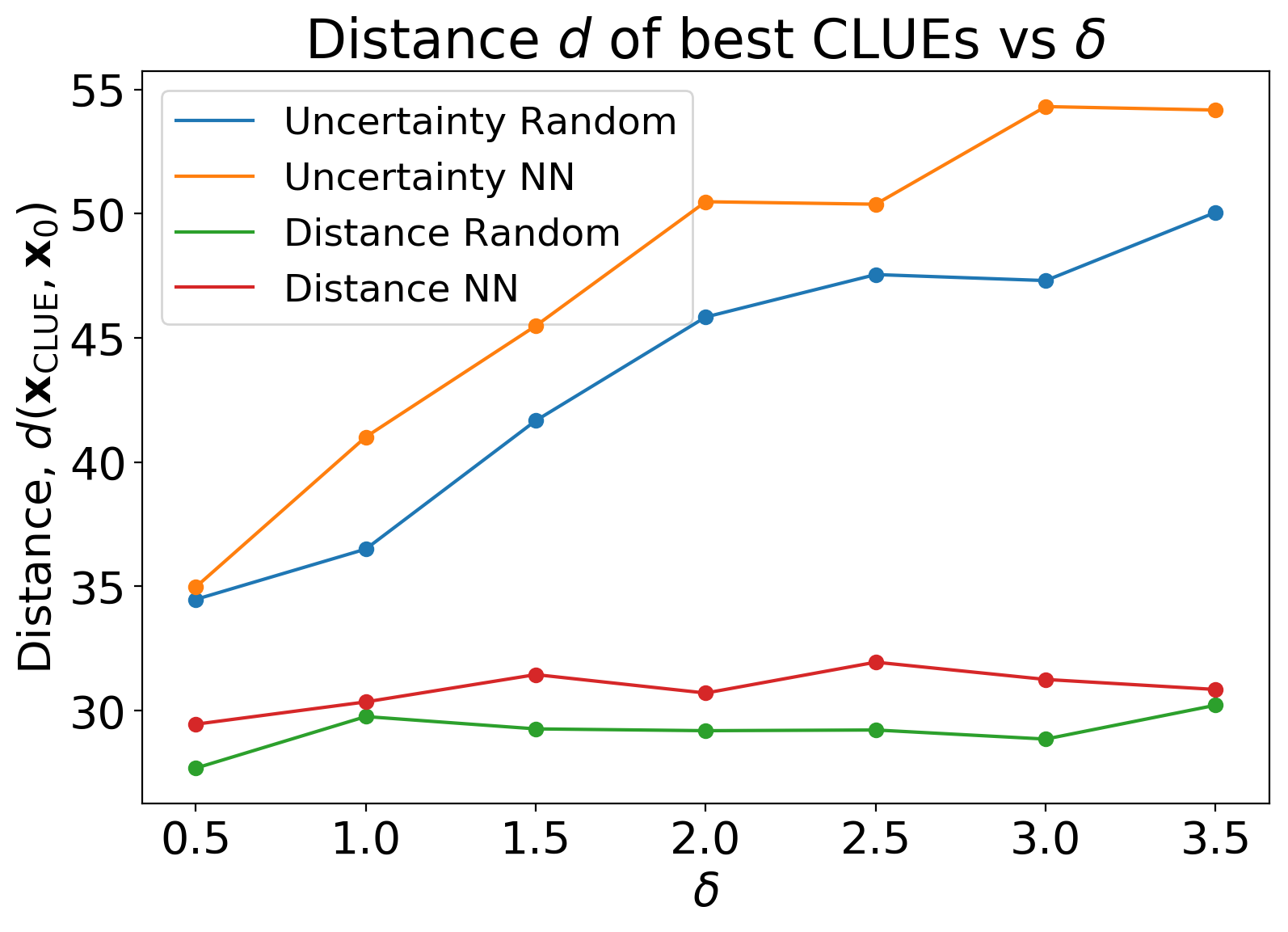}
\end{subfigure}
\caption{\small Left: Increasing the size of the $\delta$ ball yields lower uncertainty CLUEs. Right: The average distance of CLUEs from $\mathbf{x}_0$ increases with $\delta$. Note that scheme $\mathcal{S}_1$ (blue and green) outperforms scheme $\mathcal{S}_2$ (orange and red) for this dataset.}
\label{fig:uncertdistmin}
\end{figure}

\vspace{4pt}
In Figure~\ref{fig:uncertdistmin}, the $\mathcal{L}_{\mathcal{H}}$ 
experiments (blue and orange) demonstrate how the best CLUEs found improve as the $\delta$ ball expands, at the cost of increased distance from the original input. The $\mathcal{L}_{\mathcal{H}+d}$ 
experiments (green and red) suggest that the $\mathcal{L}_{\mathcal{H}+d}$ objective can vastly improve performance when it comes to distance (right), at the expense of higher (but acceptable) uncertainty.

\begin{tcolorbox}[width=\textwidth, left=2pt, right=2pt, top=1pt, bottom=1pt]    
   \textbf{Takeaway 1:} as $\delta$ increases, using either loss $\mathcal{L}_{\mathcal{H}}$ or $\mathcal{L}_{\mathcal{H}+d}$, we reduce the uncertainty of our CLUEs at the expense of greater distance $d$. Loss $\mathcal{L}_{\mathcal{H}+d}$ experiences larger performance gains in the distance curves (green and red, Figure \ref{fig:uncertdistmin}, right). 
\end{tcolorbox}

\begin{figure}[H]
\begin{subfigure}{0.5\textwidth}
    \centering
    \includegraphics[scale=0.3]{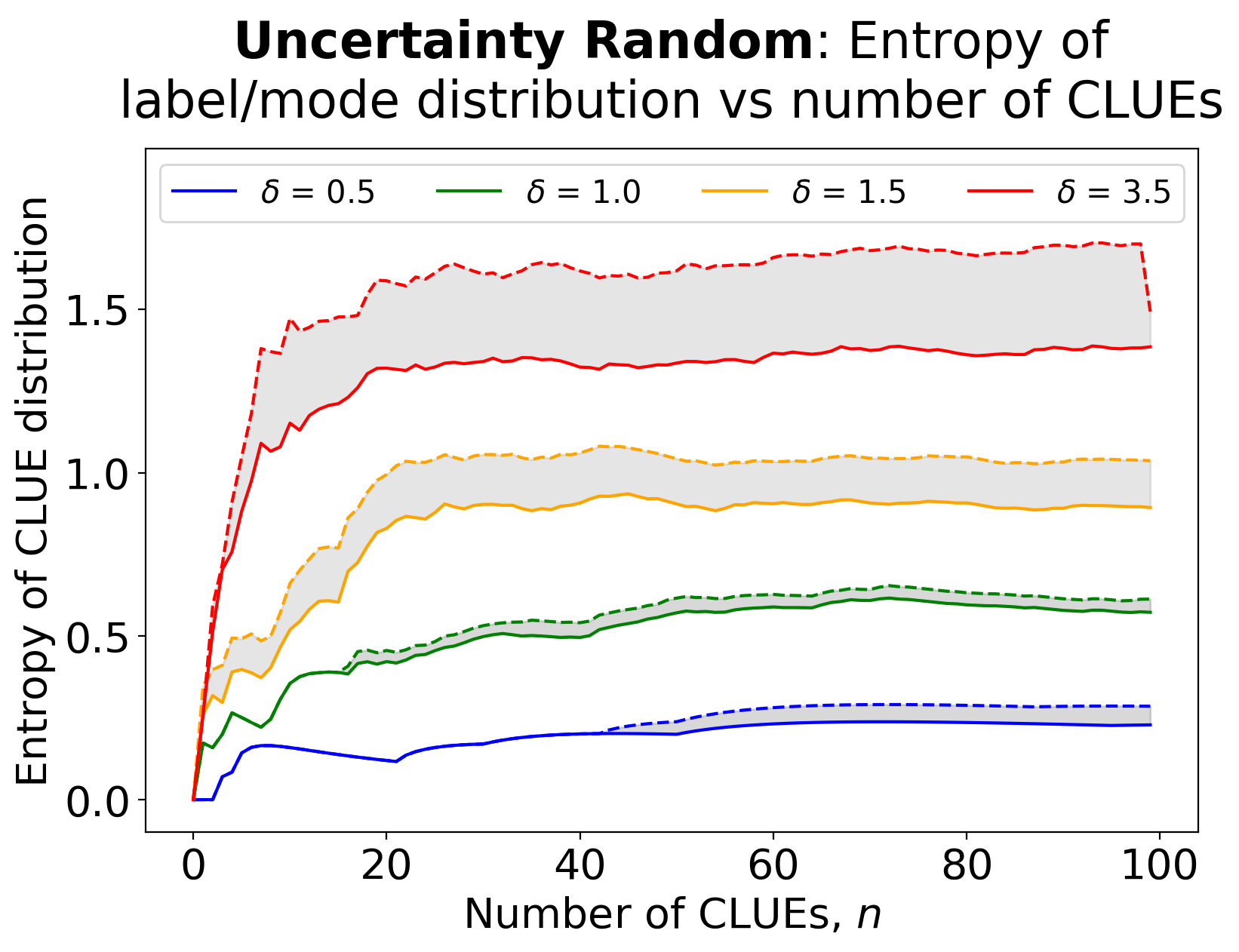}
\end{subfigure}
\begin{subfigure}{0.5\textwidth}
    \centering
    \vspace{0.3cm}
    \includegraphics[scale=0.3]{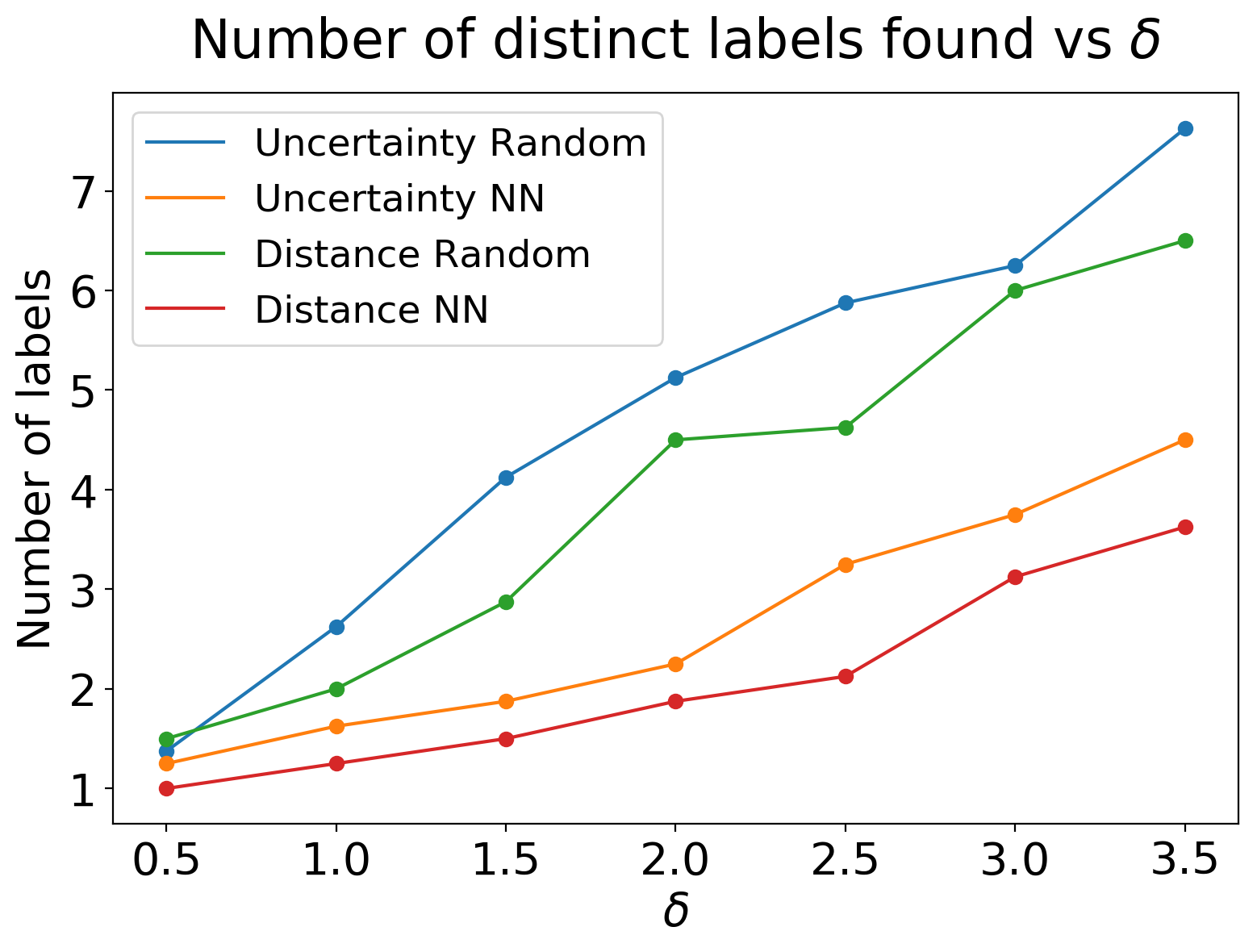}
\end{subfigure}
\caption{\small Left: Entropy of the distribution of class labels (solid) and different modes (dashed) found as number of CLUEs increases. Labels vary from 0 to 9 in MNIST whilst there exist multiple modes within each label. Observe the entropy saturating as we converge to all minima within the $\delta$ ball. Right: Average number of distinct labels found by sets of 100 CLUEs as $\delta$ increases. For small $\delta$, typically only 1 class exists (low diversity). The random search $\mathcal{S}_1$ (blue and green) achieves the greatest diversity.}
\label{fig:entropydiversity}
\end{figure}

We demonstrate that $\delta$-CLUEs are successful in converging sufficiently to all local minima within the ball, given large enough $n$ (Figure \ref{fig:entropydiversity}, left). Additionally, as the size of the $\delta$ ball increases, the random generation scheme $\mathcal{S}_1$ used in experiments \textbf{Uncertainty Random} and \textbf{Distance Random} converge to the highest numbers of diverse CLUEs (Figure \ref{fig:entropydiversity}, right, blue and green). In both loss function landscapes ($\mathcal{L}_{\mathcal{H}}$ and $\mathcal{L}_{\mathcal{H}+d}$), we obtain similarly high levels of diversity as $\delta$ increases.

\begin{tcolorbox}[width=\textwidth, left=2pt, right=2pt, top=1pt, bottom=1pt]   
\textbf{Takeaway 2:} we can achieve a diverse plethora of high quality CLUEs when it comes to both class labels and modes of change within classes, permitting a full summary of uncertainty.
\end{tcolorbox}

\begin{minipage}{0.475\textwidth}
\begin{figure}[H]
    \centering
    \includegraphics[scale=0.18]{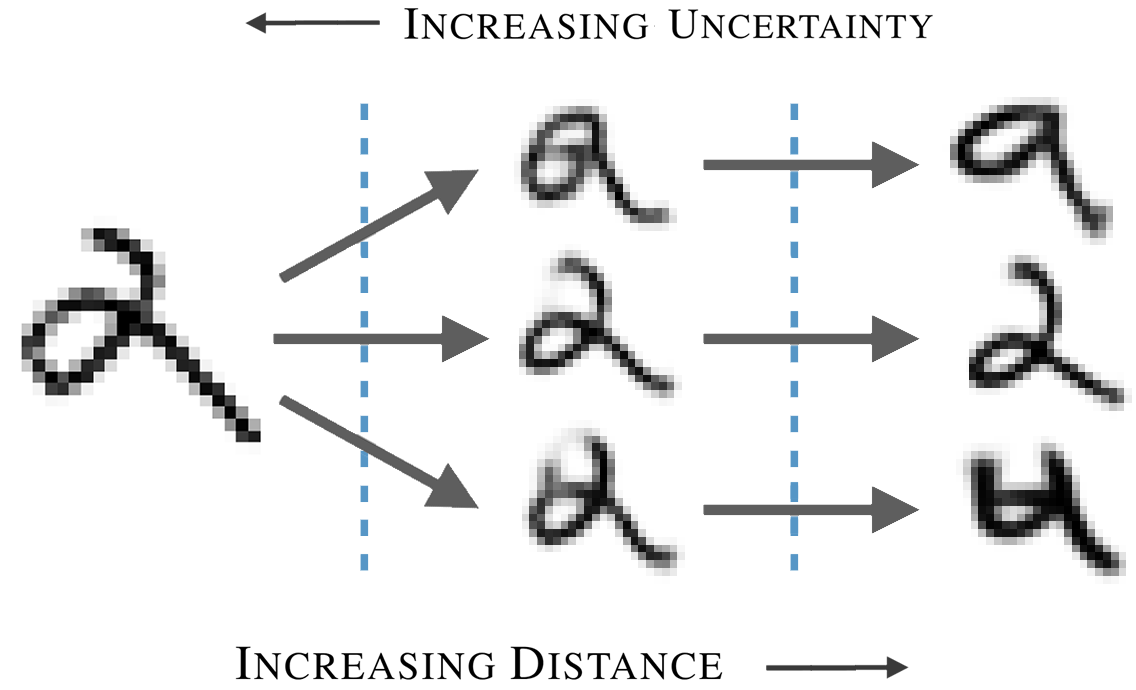}
    \caption{\small MNIST visualisation of the trade off between uncertainty $\mathcal{H}$ and distance $d$ (example of 3 diverse labels discovered by $\delta$-CLUE).}
    \label{fig:progress}
\end{figure}
\end{minipage}
\hspace{0.05\textwidth}
\begin{minipage}{0.475\textwidth}
Given a diverse set of proposed $\delta$-CLUEs (Figure \ref{fig:progress}), the performances of each class can be ranked by choosing an appropriate $\delta$ value and loss $\mathcal{L}$ for the mentioned trade offs (see Appendix \ref{appendix:labeldistribution}). Here, the digit 2 achieves lower uncertainty for a given distance, whilst the 9 and 4 require higher distances to achieve the same uncertainty. Without a $\delta$ constraint, we can move far from the original input and obtain a CLUE from any class that is certain to the BNN.
\vspace{0.2cm}
\begin{tcolorbox}[width=\textwidth, left=2pt, right=2pt, top=1pt, bottom=1pt]
\textbf{Takeaway 3:} we can produce a \textbf{label distribution} over the $\delta$-CLUEs to better summarise the diverse changes that could be made to reduce uncertainty.
\end{tcolorbox}
\end{minipage}

\section{Conclusion}

We propose $\delta$-CLUE, a method for
suggesting multiple and diverse changes to an uncertain input that (i) are local to the input and (ii) reduce the uncertainty of the input with respect to the probabilistic model. We can effectively control the trade-off between uncertainty reduction and distance by a) constraining the search within a hypersphere of radius $\delta$ and/or b) introducing a distance penalty to the objective function $\mathcal{L}(\mathbf{z})$. We demonstrate diversity in the CLUEs found on MNIST. Diversity arises via convergence to multiple class labels and to different modes of changes within these labels. 
Practitioners can use $\delta$-CLUE to understand the ambiguity of an input to a probabilistic model by suggesting a set of nearby points in the latent space of a DGM where the model is certain. For example, an uncertain $7$ might be ``close'' to a certain $7$ but also ``close'' to a certain $9$, as seen in Figure~\ref{fig:digits}.
While we manually assess mode diversity, future work could deploy a clustering algorithm for automatic assessment of various modes (i.e., different forms of the digit $7$).
As recent work considered specifying the exact level of uncertainty desired in a sample~\citep{booth2020bayes} and has considered using DGMs to find counterfactual explanations though not for uncertainty~\citep{joshi2018xgems}, we posit that leveraging DGMs to study the \textit{diversity} of plausible explanations is a promising direction to pursue. $\delta$-CLUE is just one step towards realising this goal.




\subsubsection*{Acknowledgments}
UB acknowledges support from DeepMind and the Leverhulme Trust via the Leverhulme Centre for the Future of Intelligence (CFI) and from the Mozilla Foundation.
AW acknowledges support from a Turing AI Fellowship under grant EP/V025379/1, The Alan Turing Institute under EPSRC grant EP/N510129/1 and TU/B/000074, and the Leverhulme Trust via CFI. The authors 
thank Javier Antor\'{a}n for his helpful comments and pointers.

\bibliography{iclr2021_conference}
\bibliographystyle{iclr2021_conference}

\pagebreak
\appendix
\section{Distance Metrics}
\label{appendix:distancemetrics}

In this work, we take $d_x(\mathbf{x}, \mathbf{x}_0) = \|\mathbf{x}-\mathbf{x}_0\|_1$ to encourage sparse explanations. The original CLUE paper found that for regression, $d_y(f(\mathbf{x}), f(\mathbf{x}_0))$ is mean squared error, and for classification, cross-entropy is used, noting that the best choice for $d(\cdot,\cdot)$ will be task-specific.

In some applications, these simple metrics may be insufficient, and recent work by \cite{zhang2018unreasonable} alludes to the shortcomings of even more complex distance metrics such as PSNR and SSIM. For MNIST digits (28x28 pixels), \textit{Mahanalobis distance} has been shown to be effective \citep{weinberger2009distance}, as well as other methods that achieve translation invariance \citep{grover2019mnist}.

\begin{figure}[H]
    \centering
    \includegraphics[scale=0.44]{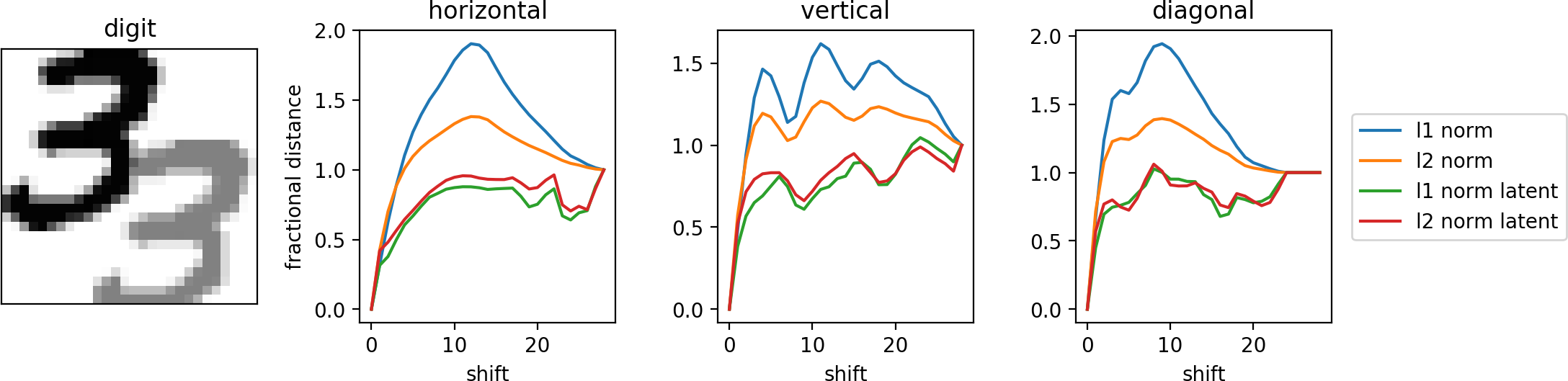}
    \caption{\small We apply horizontal, vertical and diagonal translations of an MNIST digit (in both input space and latent space for both $\ell_1$ and $\ell_2$ norms). As we increase the shift (in pixels), we compute the distance between the shifted and original digits, divided by the distance between an empty image and the original (to normalise over different metrics, resulting in convergence to 1.0). For reference, the shaded digit indicates the original digit shifted diagonally by 10 pixels.}
    \label{fig:translationrobustness}
\end{figure}

For instance, the experiment in Figure \ref{fig:translationrobustness} details how simple distance norms (either in input space and latent space) lack robustness to translations of even 5 pixels.

\section{Constrained vs Unconstrained Search}
\label{appendix:constrainedvsunconstrained}

\begin{figure}[H]
    \centering
    \includegraphics[scale=0.35]{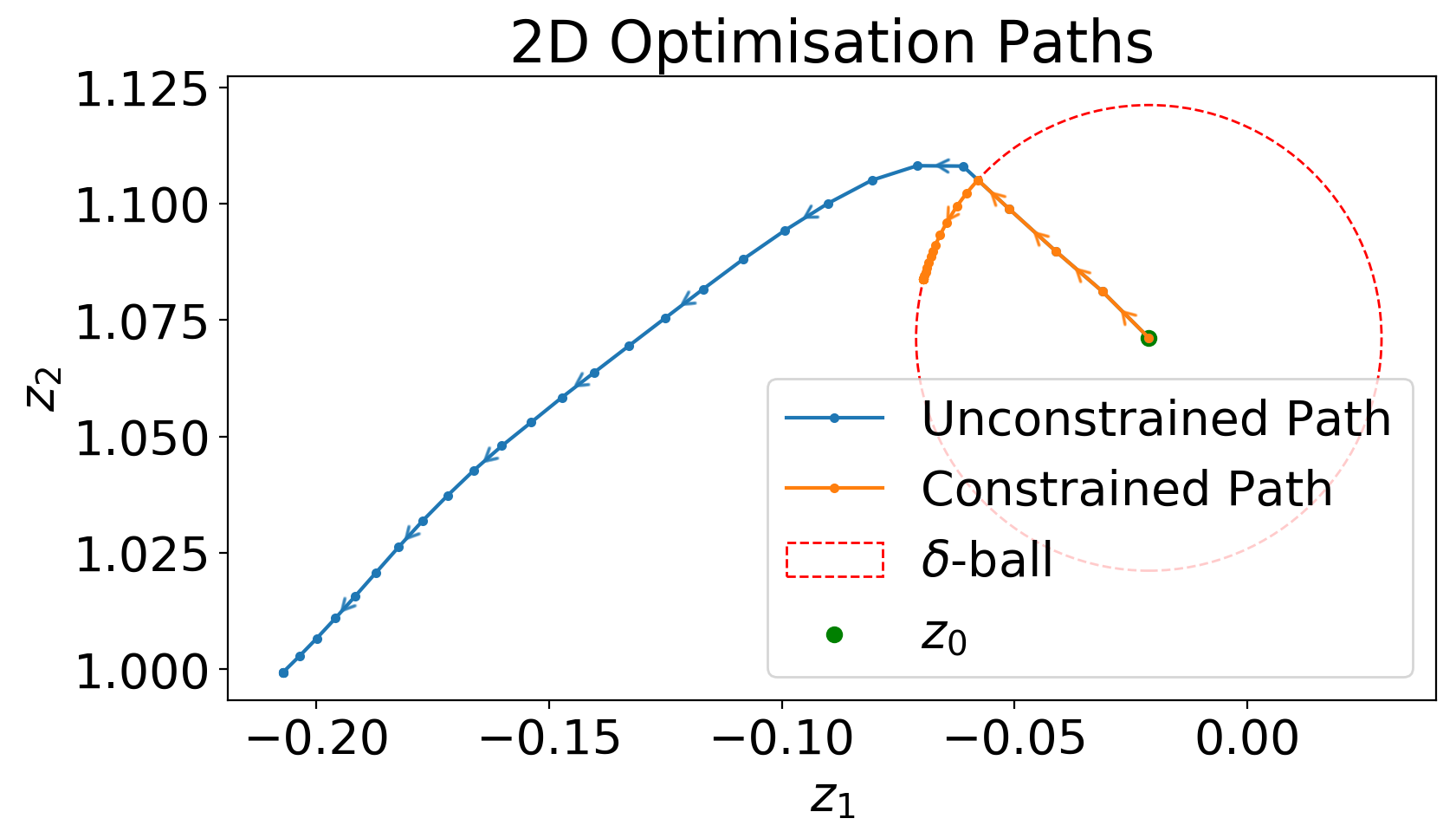}
    \caption{\small Constrained vs unconstrained gradient descents in a 2D VAE latent space $\mathcal{L}(\mathbf{z})=\mathcal{H}$. We project values outside of the $\delta$ ball onto its surface at each step of the gradient descent.}
    \label{fig:constrainedvsunconstrained}
\end{figure}

Using the $\mathcal{L}_{\mathcal{H}}=\mathcal{H}$ loss function, finding minima within the $\delta$ ball is rare for small $\delta$, and so it is necessary to use a constrained optimisation method in our experiments (Figure \ref{fig:constrainedvsunconstrained}), to avoid all solutions lying outside of the ball and being rejected.

\begin{figure}[H]
\begin{subfigure}{0.5\textwidth}
    \centering
    \includegraphics[scale=0.28]{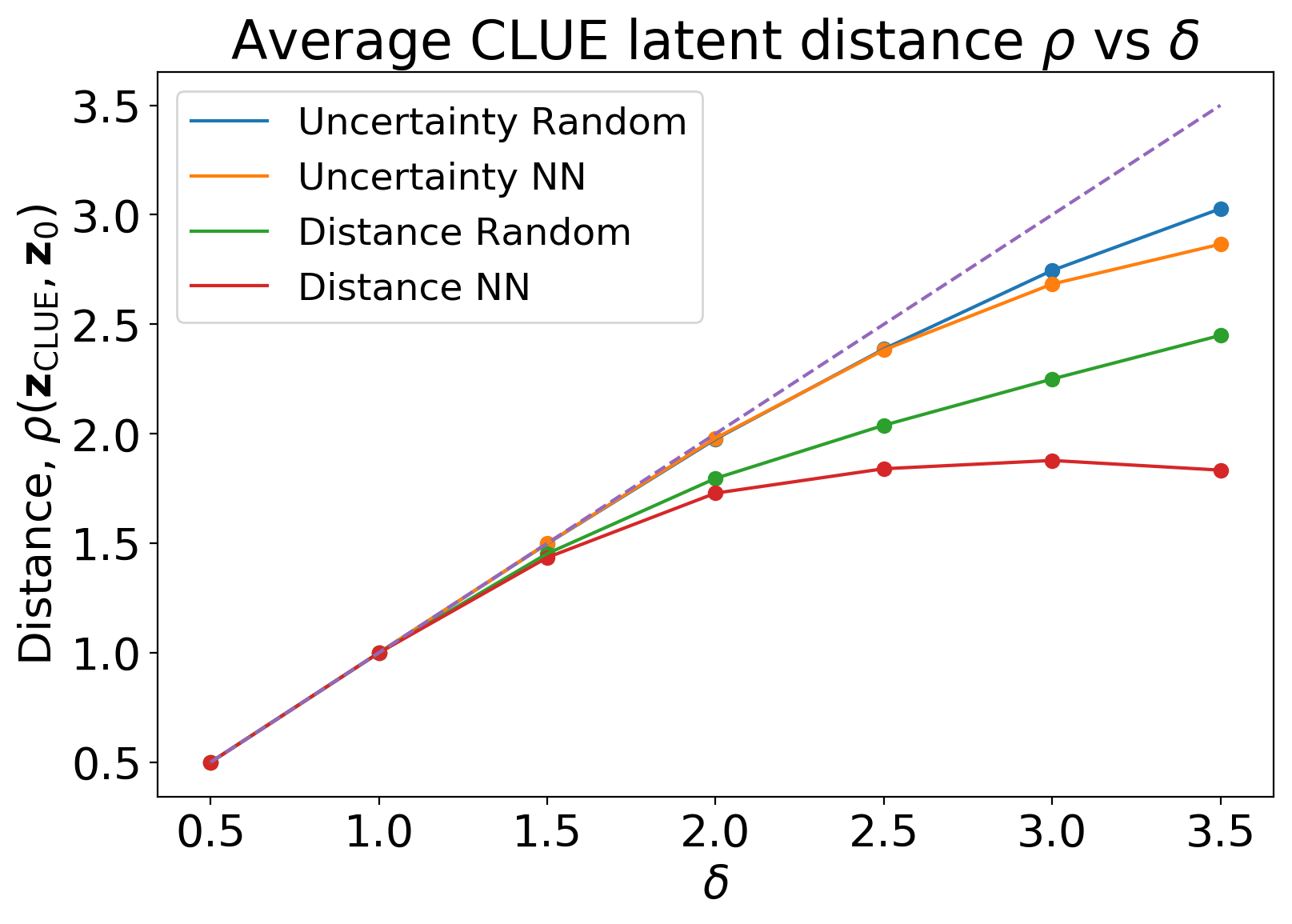}
\end{subfigure}
\begin{subfigure}{0.5\textwidth}
    \centering
    \includegraphics[scale=0.28]{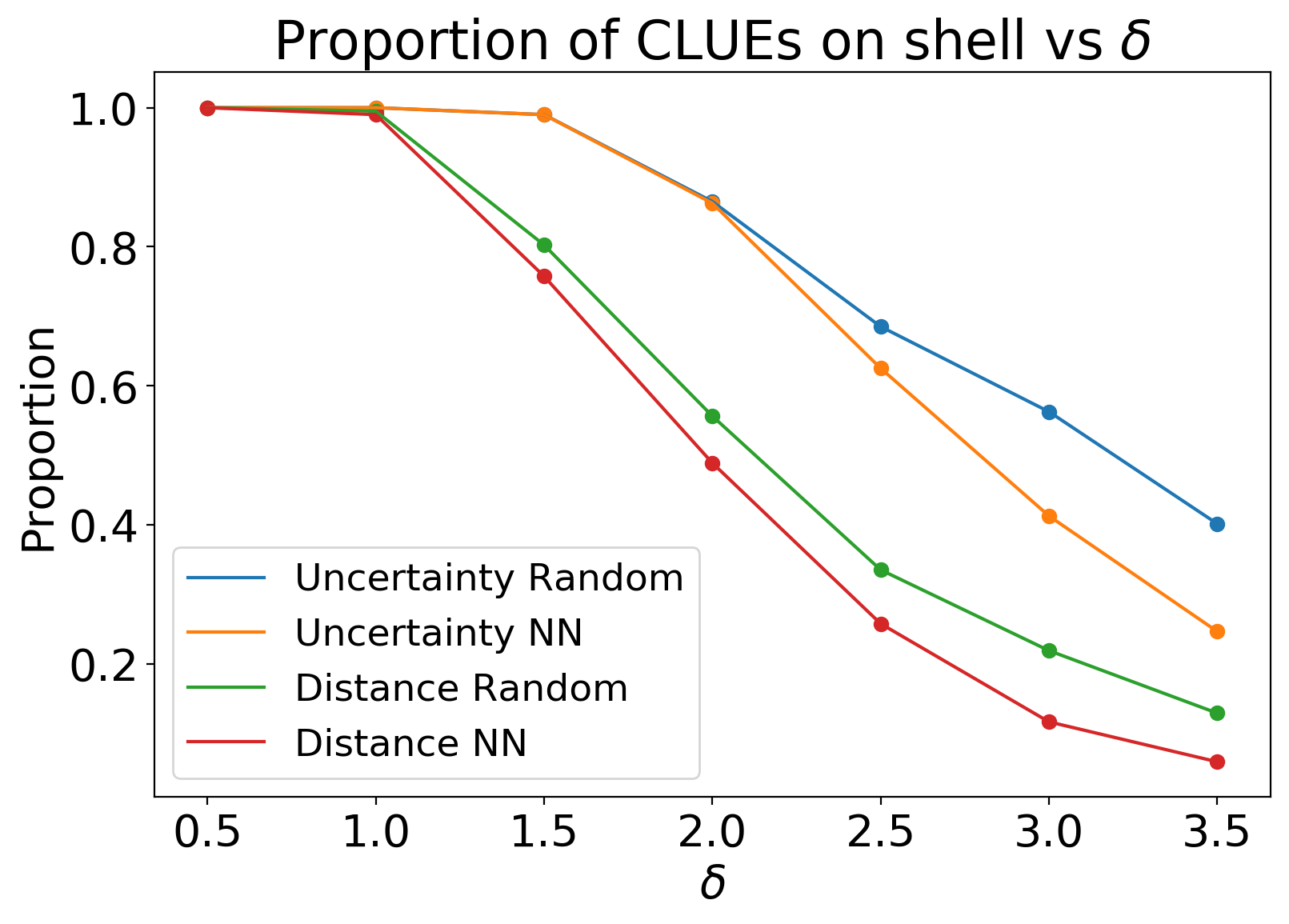}
\end{subfigure}
\caption{\small Justification for use of a constrained method. More solutions lie on the ball for a given $\delta$, instead of within it. Left: How the average final distance in latent space varies with $\delta$. Right: proportion of points that lie on the shell as $\delta$ increases. At small $\delta$, almost all minima lie on the shell, whereas at larger $\delta$ more lie inside.}
\label{fig:zdists}
\end{figure}

Thus, we observe in Figure \ref{fig:zdists}, right, that for small $\delta$, virtually all $\delta$-CLUEs lie on the surface of the ball. The left hand figure indicates that average latent space distances $\rho(\mathbf{z}_\mathrm{CLUE}, \mathbf{z}_0)$ lie close to the line $\delta=\delta$ (purple, dashed), with the distance weighted loss $\mathcal{L}_{\mathcal{H}+d}=\mathcal{H}+d$ producing more nearby $\delta$-CLUEs, as expected. In either case, the effect of the constraint weakens for larger $\delta$, as more minima exist within the ball instead of on it. Depending on user preference, the optimal $\delta$ value represents the trade off between the loss of uncertainty and the distance from the original input.

\textbf{As suggested in the main text}, there may exist methods to determine $\delta$ pre-experimentation; the distribution of training data in the latent space of the DGM could potentially uncover relationships between uncertainty and distance, both for individual inputs and on average. For instance, we might search in latent space for the distance to nearest neighbours within each class to determine $\delta$. In many cases, it could be useful to provide a summary of counterfactuals at various distances and uncertainties, making a range of $\delta$ values more appropriate.

\section{Initialisation Schemes $\mathcal{S}_i$}
\label{appendix:schemes}

This appendix details the initialisation schemes $\mathcal{S}_i$ that are used to generate start points for the algorithm. While some schemes may appear preferential in 2 dimensions, the manner at which these scale up to higher dimensions means that we could require an infeasible number of initialisations to cover the appropriate landscape, and so deterministic schemes such as a path towards nearest neighbours within each class ($\mathcal{S}_2$), or a gradient descent into predictions within each class ($\mathcal{S}_5$) might be desirable. The following mathematical analysis applies to an $\ell_2$-norm $\rho(\mathbf{z}, \mathbf{z}_0)=\|\mathbf{z}-\mathbf{z}_0\|_2$:

\[\mathcal{S}_1: \rho(\mathbf{z}, \mathbf{z}_0)\sim \mathcal{U}(0, \delta)\implies\E[\mathbf{\rho(\mathbf{z}, \mathbf{z}_0)}]=\frac{\delta}{2}\ \text{ (pick a random radial direction)}\]
\[\mathcal{S}_3:\rho(\mathbf{z}, \mathbf{z}_0)\sim \mathcal{N}\left(0, \frac{\delta^2}{4}\right)\ \text{s.t. } 0\leq\rho(\mathbf{z}, \mathbf{z}_0)\leq\delta\ \text{ (pick a random radial direction)}\]
\[\mathcal{S}_4:[\mathbf{z}-\mathbf{z}_0]_i\sim \mathcal{U}\left(-\delta,\delta\right)\ \text{s.t. } \rho(\mathbf{z}, \mathbf{z}_0)\leq\delta\]

\begin{figure}[H]
\begin{subfigure}{0.33\textwidth}
    \centering
    \includegraphics[scale=0.26]{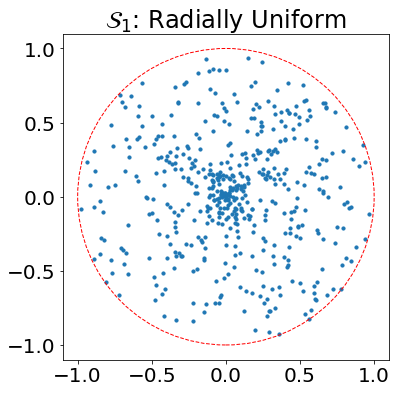}
\end{subfigure}
\begin{subfigure}{0.33\textwidth}
    \centering
    \includegraphics[scale=0.26]{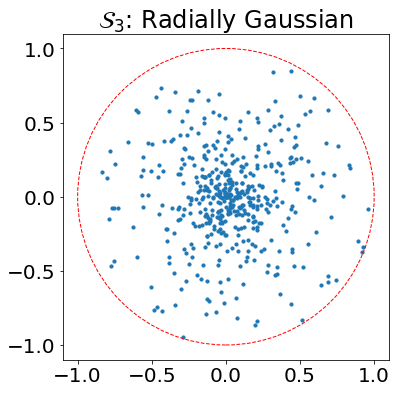}
\end{subfigure}
\begin{subfigure}{0.33\textwidth}
    \centering
    \includegraphics[scale=0.26]{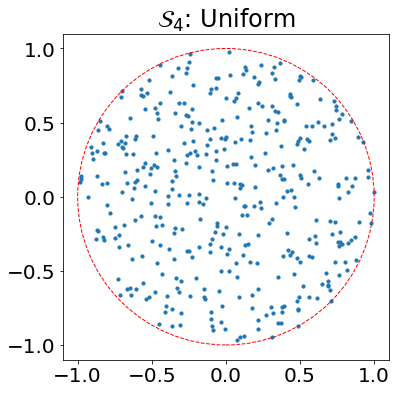}
\end{subfigure}
\caption{\small Random generation schemes $\mathcal{S}_1$, $\mathcal{S}_3$ and $\mathcal{S}_4$ depicted in 2D space. In Schemes $\mathcal{S}_3$ and $\mathcal{S}_4$ we reject samples outside of the $\delta$ ball (where $\rho(\mathbf{z}, \mathbf{z}_0)>\delta$). Future schemes may generate within a sub-ball that is smaller than the ball with which we constrain, though this may only be effective in specific latent landscapes.}
\label{fig:S}
\end{figure}

We propose two potential deterministic schemes, that may outperform a random scheme when a) the latent dimension is large, b) $\delta$ becomes very large, c) we impose a larger distance weight in the objective function or d) we change datasets. Here $\mathbf{z}_i$ represents the starting point for explanation $i$, $n$ is the total number of explanations (both used in Algorithm 1), $Y$ represents the total number of class labels $y$, and $j\in\mathbb{Z}^+$. This produces a total of $Y\times j_{\text{max}}=Y\times\left\lfloor{\frac{n}{Y}}\right\rfloor=n$ explanations if $Y|n$.

\[\mathcal{S}_2:\mathbf{z}_i = \mathbf{z}_0+\delta \times\frac{j}{m}\times \frac{\mathbf{z}_y-\mathbf{z}_0}{\rho(\mathbf{z}_y,\mathbf{z}_0)}\ \forall y\]
\[\mathcal{S}_5: \mathbf{z}_i = \mathbf{z}_0+\mathbf{s}_{yj}\ \forall y\]
\[\text{where }1\leq j\leq m\ \text{and }m=\left\lfloor{\frac{n}{Y}}\right\rfloor\]

where, for the $\mathcal{S}_5$ scheme, $\mathbf{s}_{yj}$ is defined along a path from $\mathbf{z}_0$ to a radius $\delta$, where at all points the direction of $\mathbf{s}$ is $\nabla_\mathbf{z}p(\text{class}(\mathbf{z})=y)$, and $\frac{j}{m}$ is defined as the fraction travelled along that path.

\begin{figure}[H]
    \centering
    \includegraphics[scale=0.45]{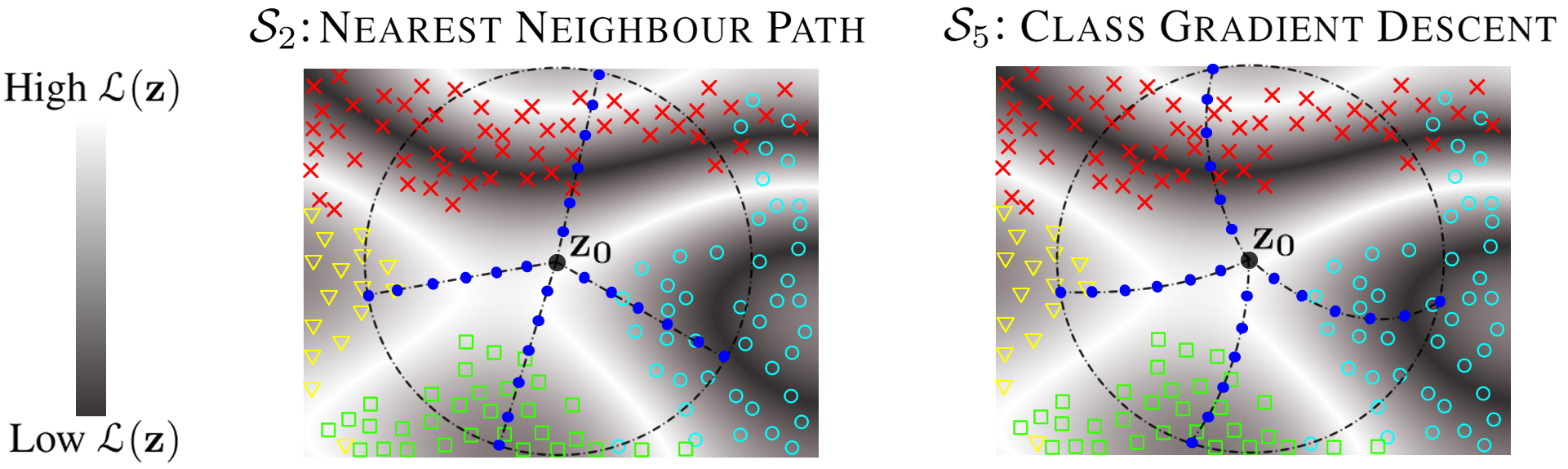}
    \caption{\small Left: Scheme $\mathcal{S}_2$, nearest neighbour path, searches for the nearest low uncertainty points in training data for each class, before initialising starting points fractionally on the path towards said neighbour. Right: Scheme $\mathcal{S}_5$ performs a gradient descent in the prediction space of the BNN, towards maximising the probability of each class. It too initialises starting points along said path.}
    \label{fig:SDeterministic}
\end{figure}

A series of modifications to these schemes may improve their performance:

\begin{itemize}
    \item Generating within small regions around each of the points along the path (in $\mathcal{S}_2$ and $\mathcal{S}_5$).
    \item Performing a series of further subsearches in latent space around each of the best $\delta$-CLUEs under a particular scheme.
    \item Combining $\delta$-CLUEs from multiple methods to achieve greater diversity.
\end{itemize}

\section{Further MNIST $\delta$-CLUE Analysis}
\label{appendix:analysis}

\begin{figure}[H]
\begin{subfigure}{0.5\textwidth}
    \centering
    \includegraphics[scale=0.28]{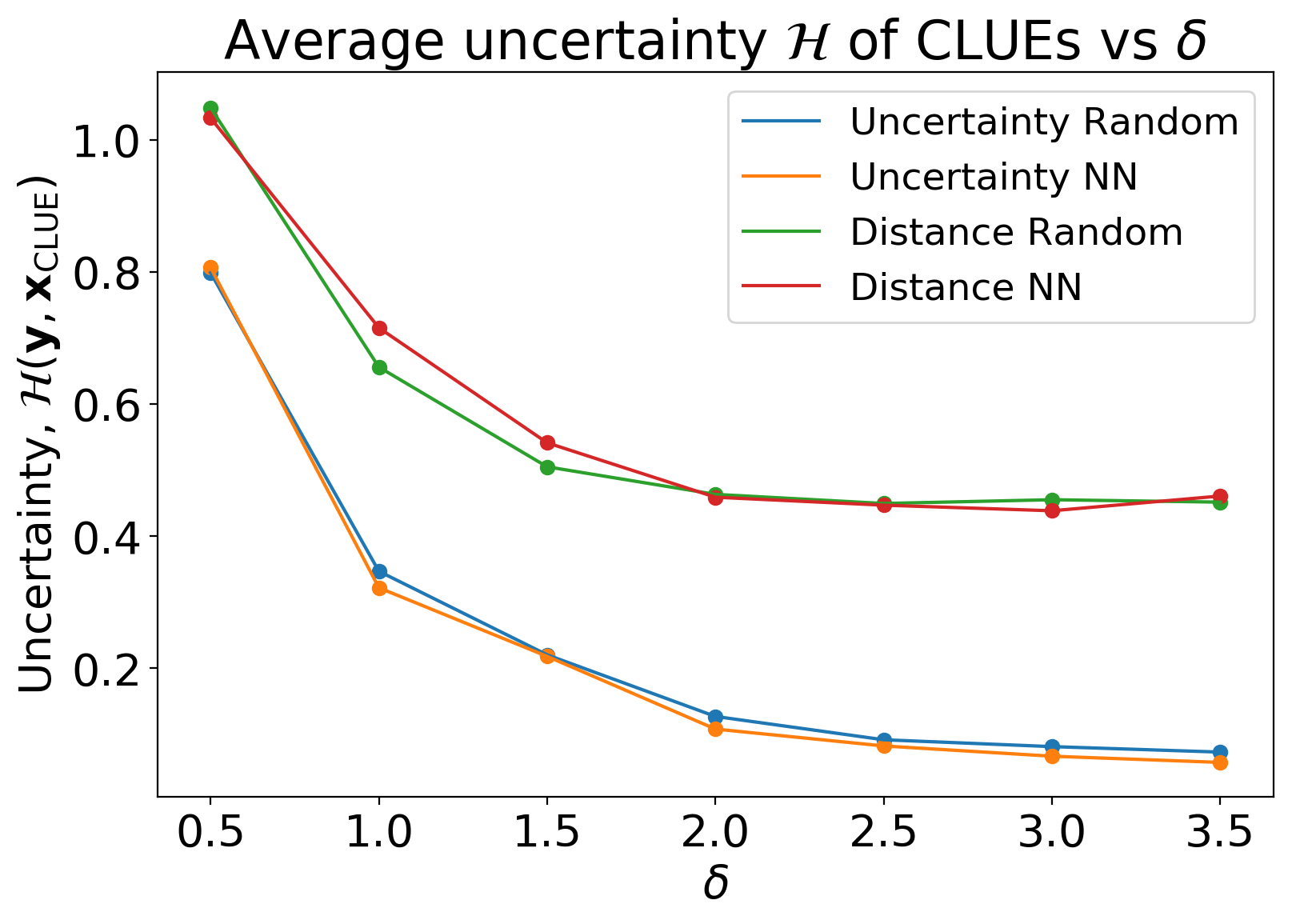}
\end{subfigure}
\begin{subfigure}{0.5\textwidth}
    \centering
    \includegraphics[scale=0.28]{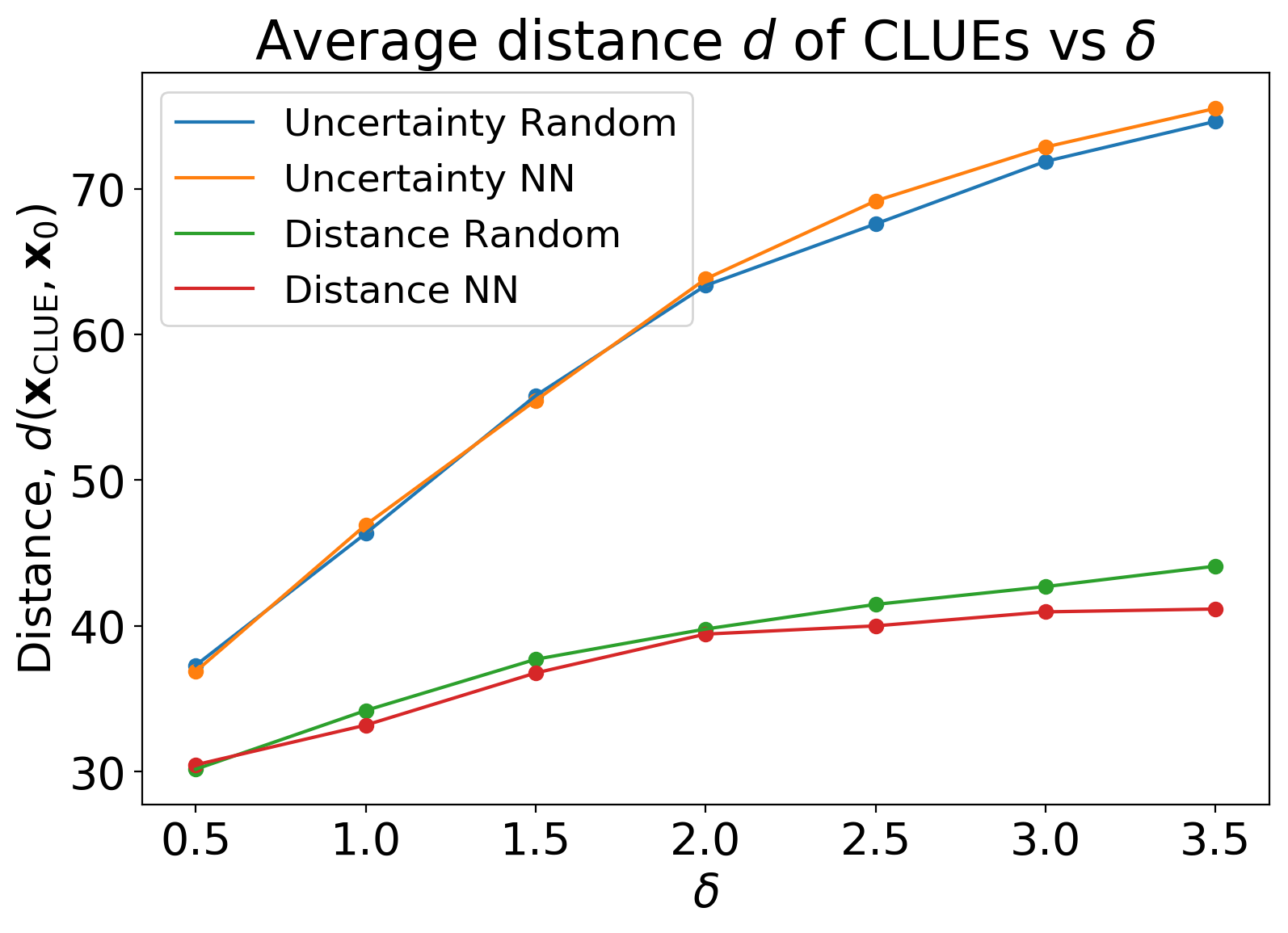}
\end{subfigure}
\caption{\small In Figure \ref{fig:uncertdistmin} of the main text, we plot the best (minimum) uncertainties/distances of the $\delta$-CLUEs. Here, we reproduce the plot for average uncertainties/distances and observe that it follows similar trends, shifted vertically, with higher disparity between the $\mathcal{L}_{\mathcal{H}}$ and $\mathcal{L}_{\mathcal{H}+d}$ loss functions.}
\label{fig:uncertdistavg}
\end{figure}

\begin{figure}[H]
\begin{subfigure}{0.5\textwidth}
    \centering
    \includegraphics[scale=0.28]{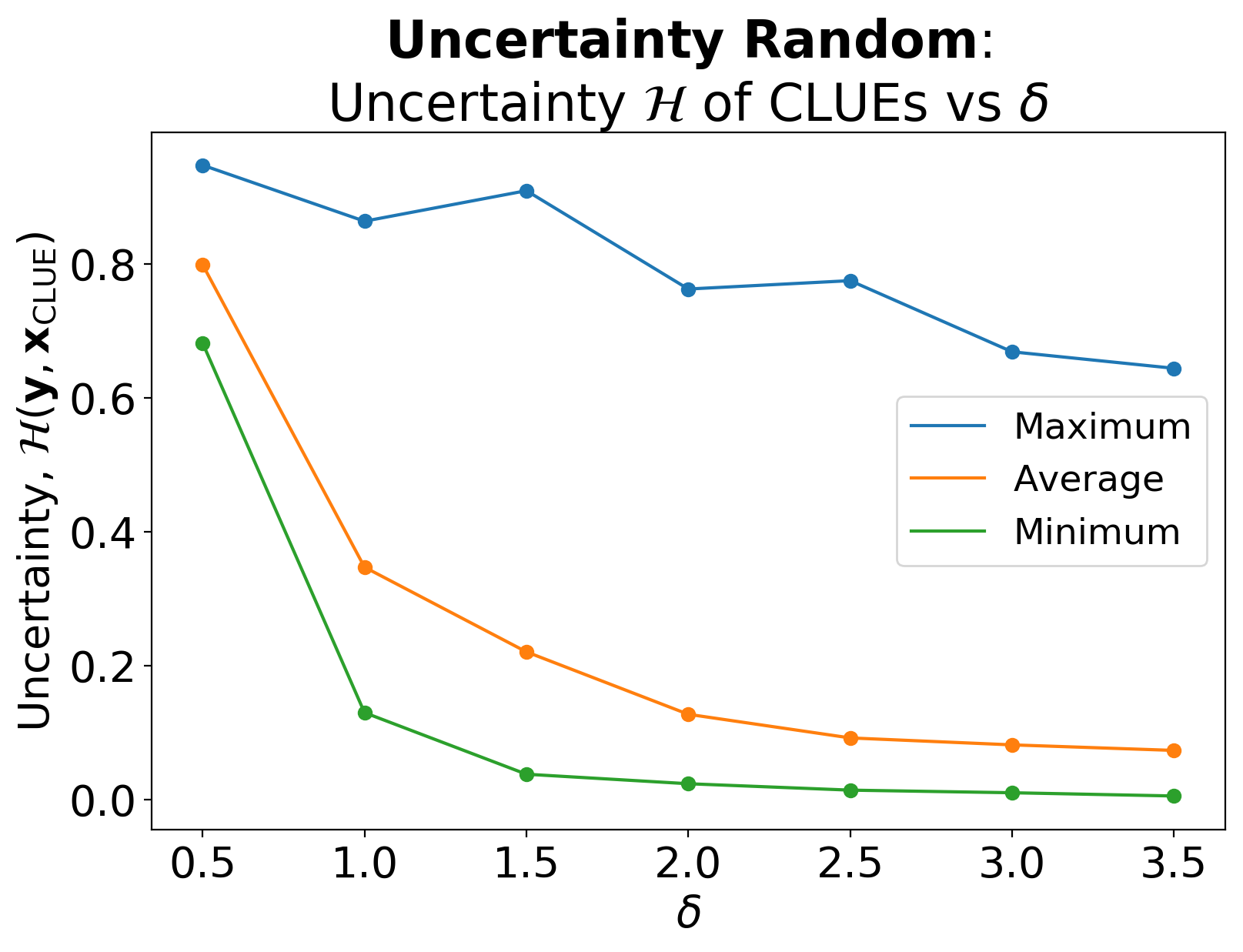}
\end{subfigure}
\begin{subfigure}{0.5\textwidth}
    \centering
    \includegraphics[scale=0.28]{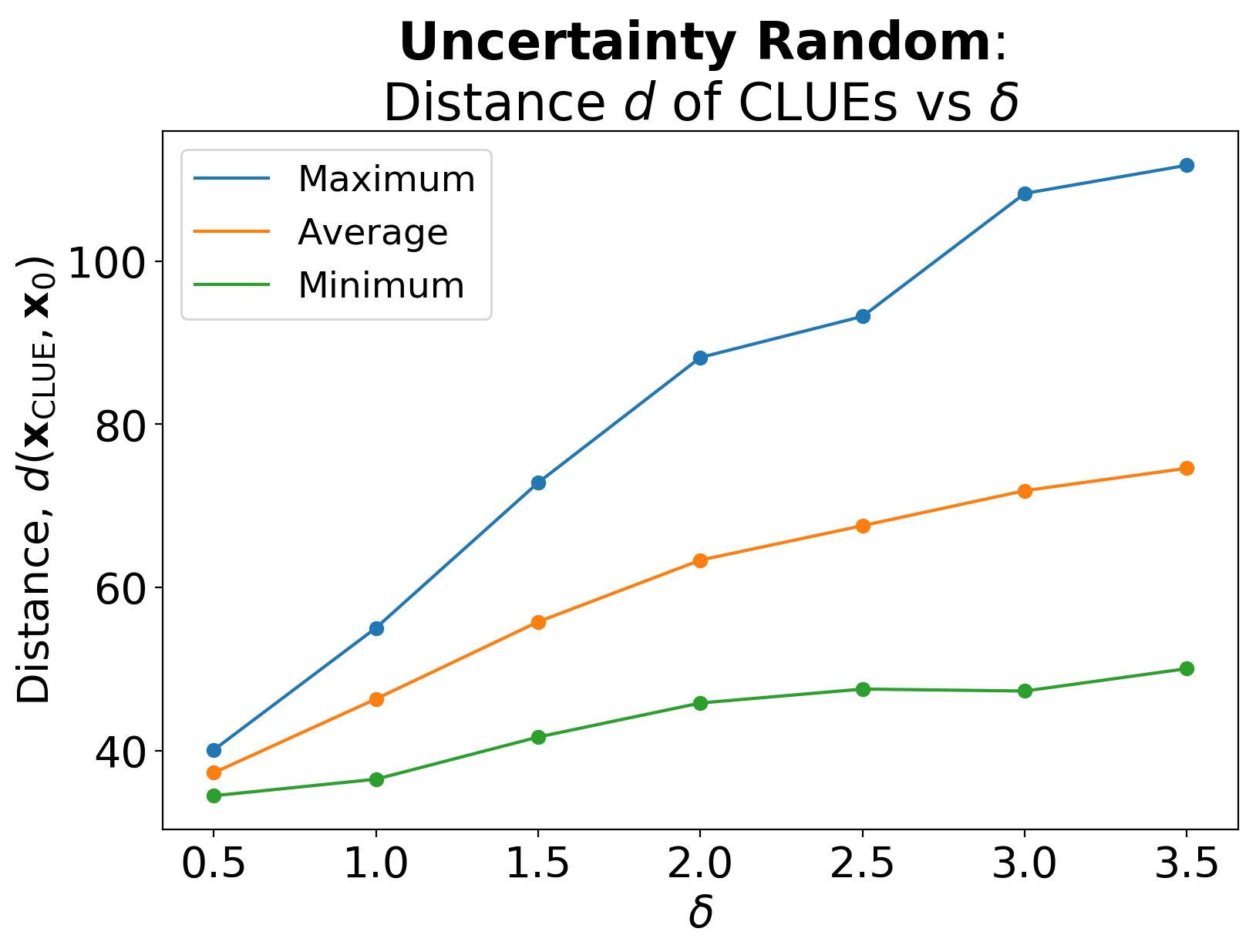}
\end{subfigure}
\caption{\small We reproduce Figure \ref{fig:uncertdistmin} for the \textbf{Uncertainty Random} experiment ($\mathcal{L}_{\mathcal{H}}=\mathcal{H}$ and $\mathcal{S}_1$), plotting the minimum, average and maximum values found in the set of 100 $\delta$-CLUEs averaged over 8 uncertain inputs.}
\label{fig:uncertdistminmax}
\end{figure}

\begin{figure}[H]
    \centering
    \includegraphics[scale=0.3]{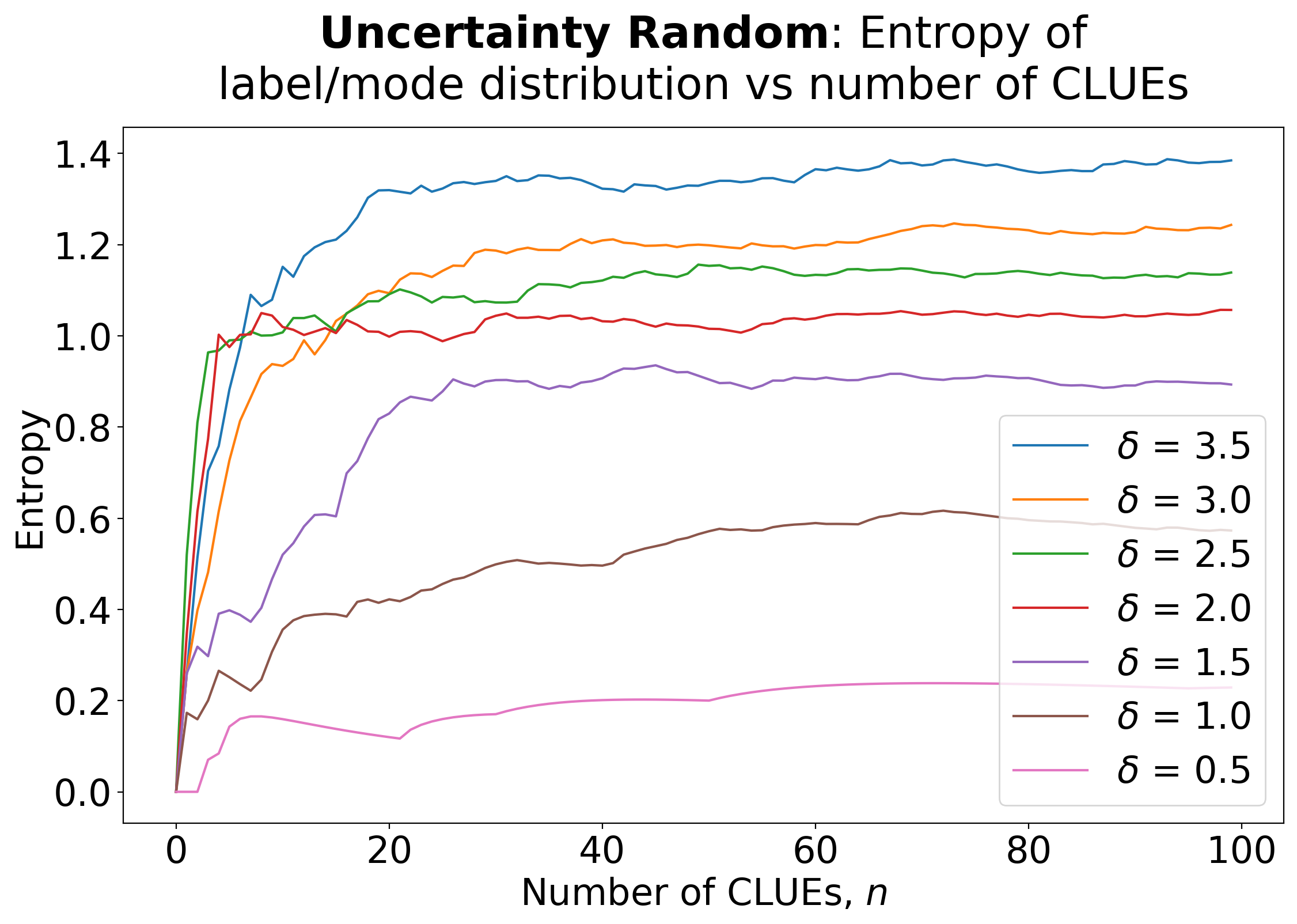}
    \caption{\small A more refined plot of Figure \ref{fig:entropydiversity}, left, to answer the question: ``How many times must we run $\delta$-CLUE in order to saturate the entropy of the label distribution of the $\delta$-CLUEs found?''.}
    \label{fig:labelentropy}
\end{figure}

For an uncertain input $\mathbf{x}_0$, we generate 100 $\delta$-CLUEs and compute the minimum, average and maximum uncertainties/distances from this set, before averaging this over 8 different uncertain inputs. Repeating this over several $\delta$ values produces Figures \ref{fig:uncertdistavg} through \ref{fig:labelentropy}.

Special consideration should be taken in selecting the best method to assess a set of 100 $\delta$-CLUEs: the minimum/average uncertainty/distance $\delta$-CLUEs could be selected, or some form of submodular selection algorithm could be deployed on the set. Figure \ref{fig:uncertdistminmax} shows the variance in performance of $\delta$-CLUEs; the worst $\delta$-CLUEs converge to high uncertainties and high distances that are too undesirable (the selection of $\delta$-CLUEs is then a non-trivial problem to solve, and in our analysis we simply select the best cost $\delta$-CLUE for each CLUE, where cost is a combination of uncertainty and distance).

\begin{figure}[H]
    \centering
    \includegraphics[scale=0.25]{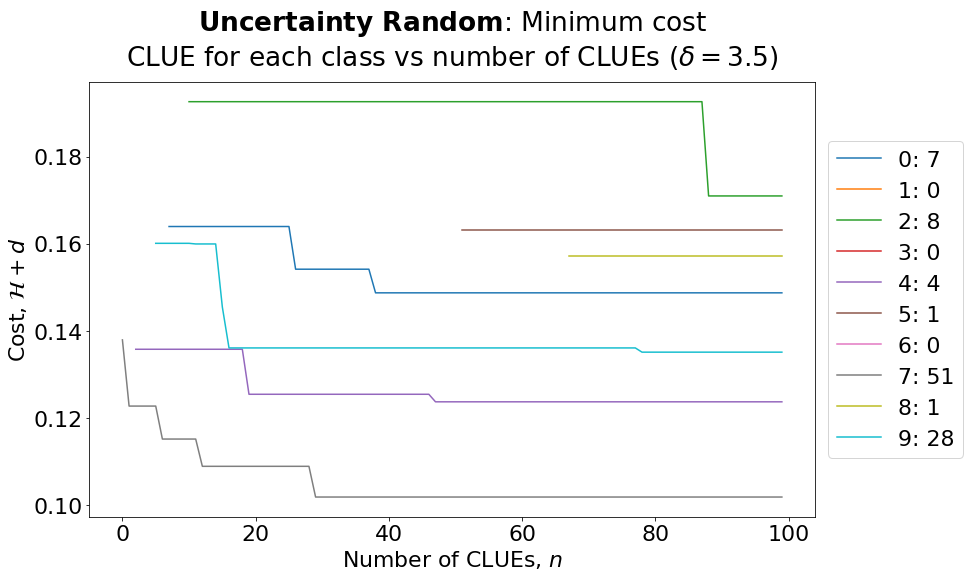}
    \caption{\small For a single uncertain input $\mathbf{x}_0$, we generate $n$ $\delta$-CLUEs and observe how the minimum cost (a combination of uncertainty and distance) of $\delta$-CLUEs for each class converges. Legend shows class labels 0 to 9, and the final number of each discovered by $\delta$-CLUE (summing to 100).}
    \label{fig:costprogression}
\end{figure}

In Figure \ref{fig:costprogression}, the late convergence of class 2 (green) and the lack of 1s, 3s and 6s suggests that $n>100$ is required, although under computational constraints $n=100$ yields good quality CLUEs for the prominent classes (7 and 9).

\begin{figure}[H]
    \centering
    \includegraphics[scale=0.18]{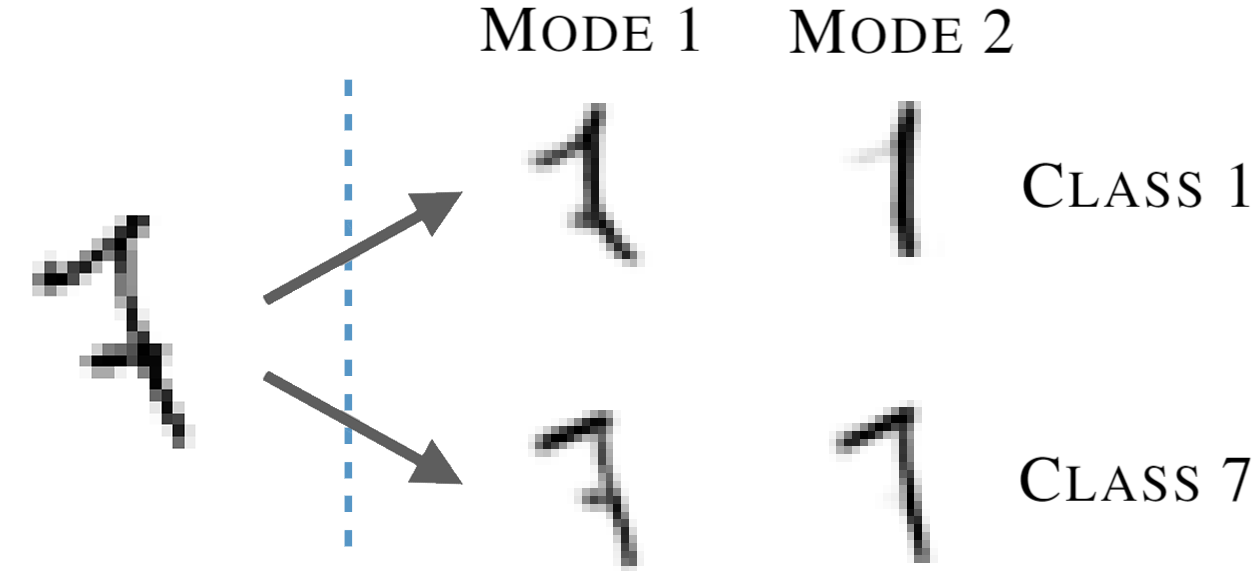}
    \caption{\small MNIST: 10 class labels exist (0 to 9), whereas an undefined number of modes within each class also exist. These modes are counted manually in this paper.}
    \label{fig:modes}
\end{figure}

Figure \ref{fig:modes} demonstrates how convergence of the $\delta$-CLUE set is a function, not only of the class labels found, but also of the different mode changes that result within each class (alternative forms of each label). In the main text (Figure \ref{fig:entropydiversity}), we count manually the mode changes within each class; in future, clustering algorithms such as Gaussian Mixture Models could be deployed to automatically assess these. The concept of modes is important when a low number of classes exists, such as in binary classification tasks, where we may require multiple ways of answering the question: ``what possible mode change could an end user make to modify their classification from a no to a yes?''.

\section{Computing a Label Distribution from $\delta$-CLUEs}
\label{appendix:labeldistribution}

This final appendix addresses the task of computing a label distribution from a set of $\delta$-CLUEs, as suggested by takeaway 3 of the main text. We use $\delta=3.5$ and analyse one uncertain input $\mathbf{x}_0$ under the experiment \textbf{Distance Random} where $\mathcal{L}_{\mathcal{H}+d}=\mathcal{H}+d$ and $\mathcal{S}_1$ are used.

\begin{figure}[H]
\begin{subfigure}{0.2\textwidth}
    \centering
    \includegraphics[scale=0.33]{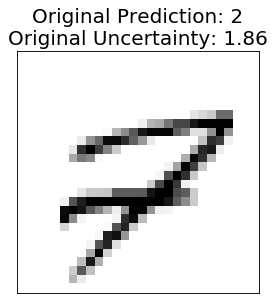}
\end{subfigure}
\begin{subfigure}{0.4\textwidth}
    \centering
    \vspace{0.6cm}
    \includegraphics[scale=0.23]{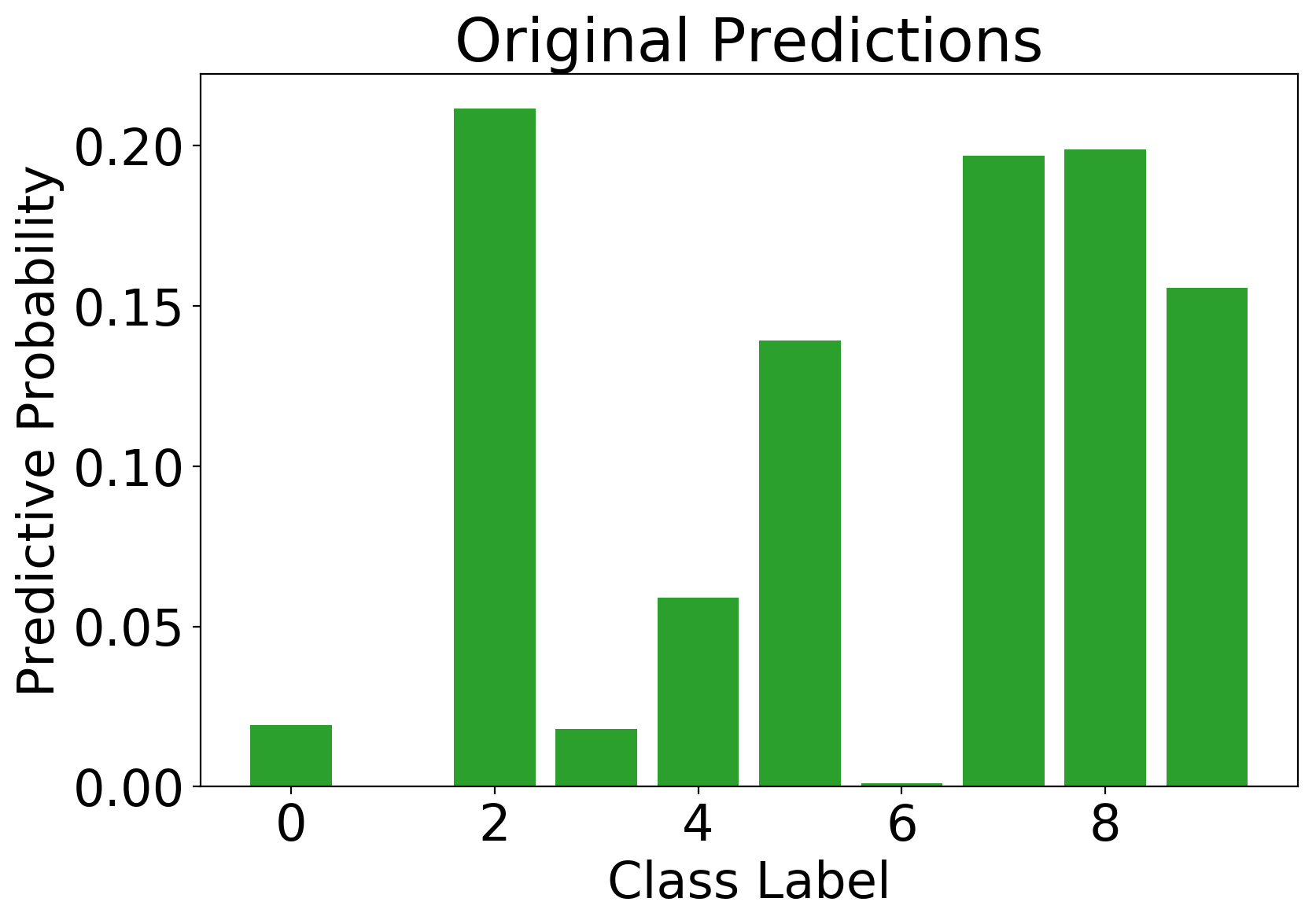}
\end{subfigure}
\begin{subfigure}{0.4\textwidth}
    \centering
    \vspace{0.6cm}
    \includegraphics[scale=0.23]{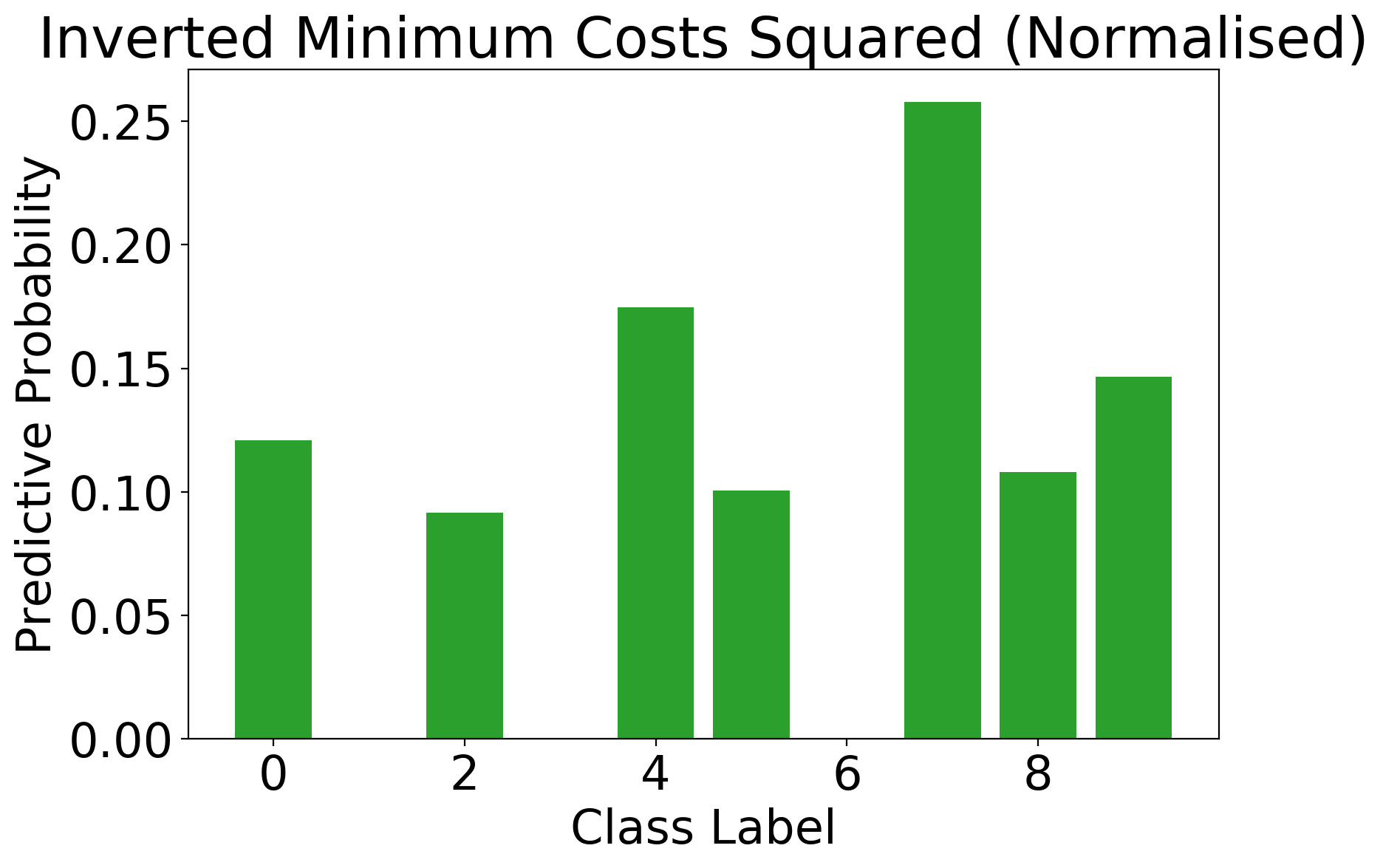}
\end{subfigure}
\caption{\small Left: An original uncertain input that is incorrectly classified. Centre: The original predictions from the BNN. Right: The new \textbf{label distribution} based off of the $\delta$-CLUEs found.}
\label{fig:labeldistribution}
\end{figure}

\begin{figure}[H]
\begin{subfigure}{0.33\textwidth}
    \centering
    \includegraphics[scale=0.2]{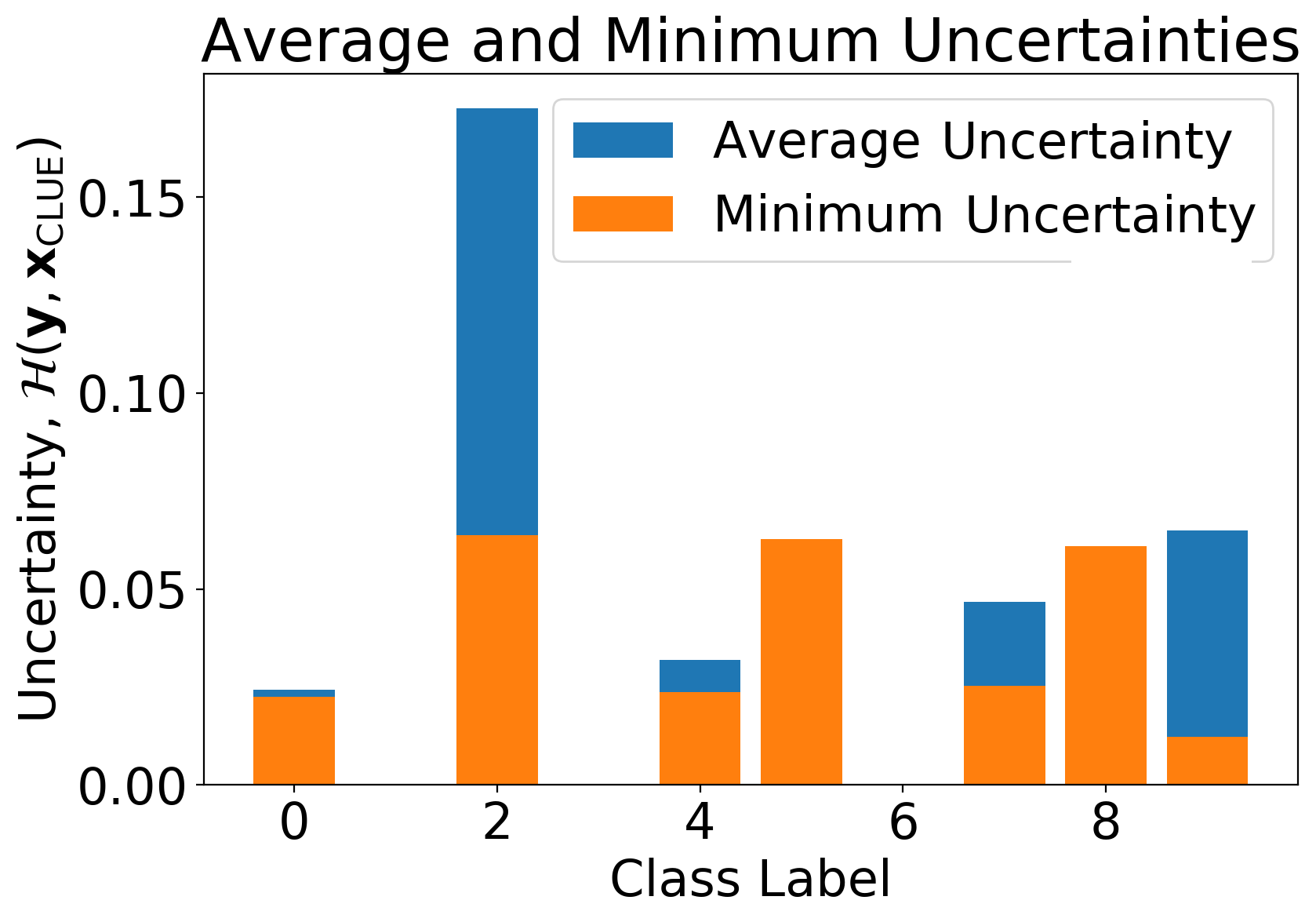}
\end{subfigure}
\begin{subfigure}{0.33\textwidth}
    \centering
    \includegraphics[scale=0.2]{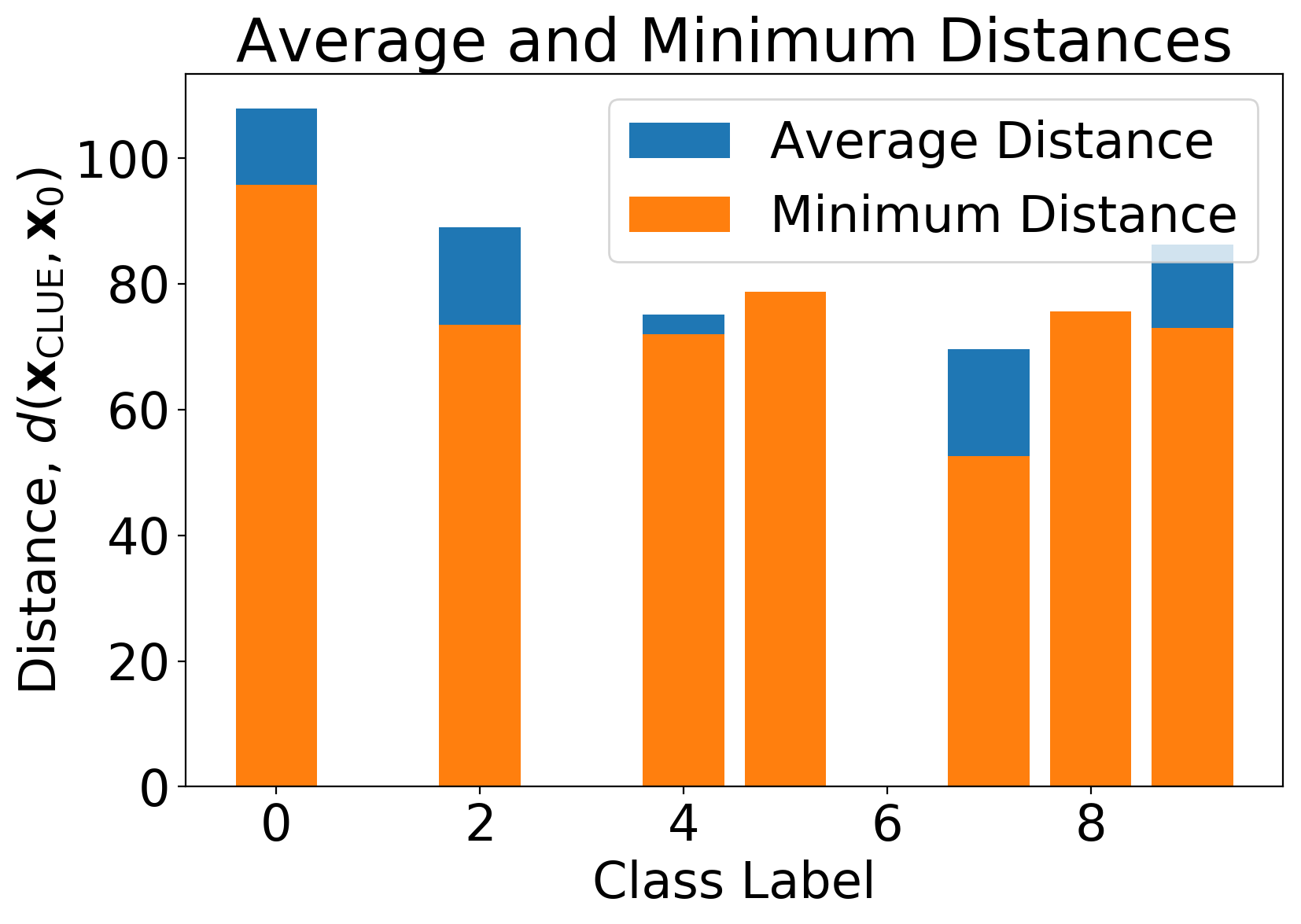}
\end{subfigure}
\begin{subfigure}{0.33\textwidth}
    \centering
    \includegraphics[scale=0.2]{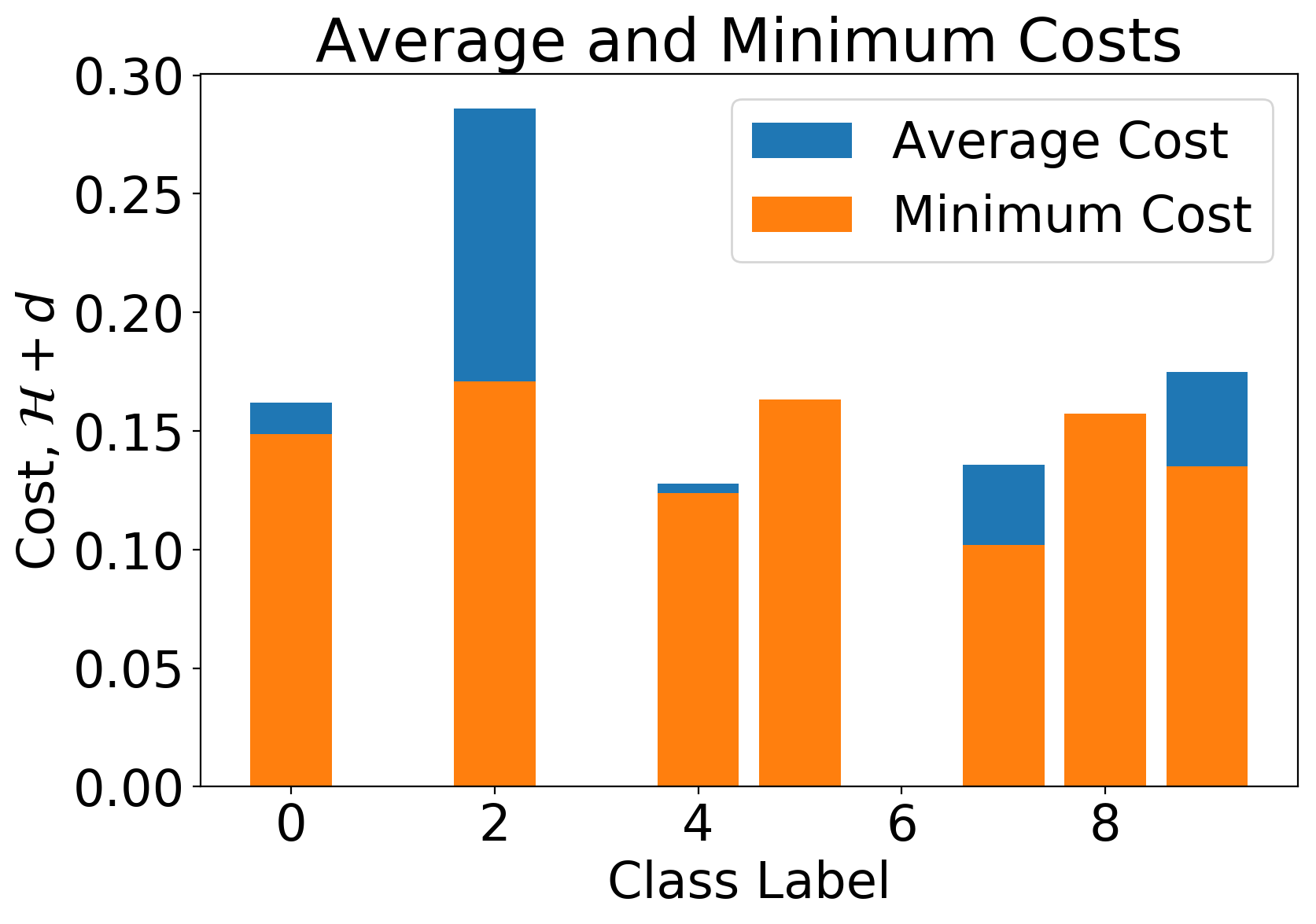}
\end{subfigure}
\caption{\small Left: Average and minimum uncertainties $\mathcal{H}$ for each class in the $\delta$-CLUE set. Centre: Average and minimum distances $d$. Right: Average and minimum costs, where the weight $\lambda_x$ is multiplied by the distance function and added to the uncertainty.}
\label{fig:label}
\end{figure}

For (Figure \ref{fig:labeldistribution}, right), we take the minimum costs from (Figure \ref{fig:label}, right) and take the inverse square.

\begin{figure}[H]
    \centering
    \includegraphics[scale=0.7]{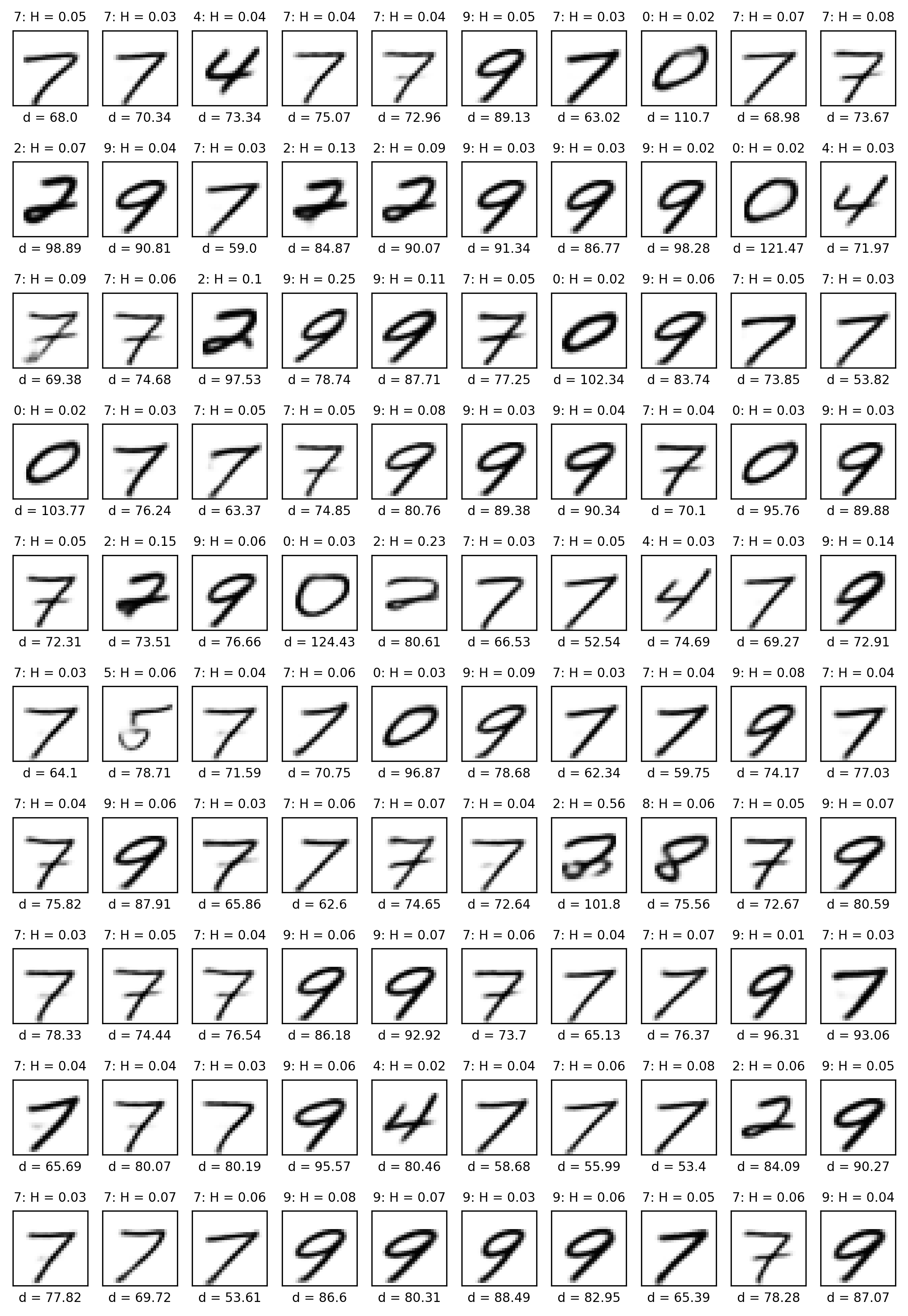}
    \caption{\small The 100 $\delta$-CLUEs yielded in this experiment (\textbf{Distance Random} with $\delta=3.5$). Above digits: Label prediction and uncertainty. Below: Distance from original in input space. Low uncertainty CLUEs may be found at the expense of a greater distance from the original input.}
    \label{fig:deltaCLUEs}
\end{figure}

\end{document}